\documentclass[10pt,journal,web]{article}
\usepackage[T1]{fontenc}
\usepackage[utf8]{inputenc}
\usepackage[english]{babel}

\usepackage{amsmath,amssymb,amsfonts}
\usepackage{amsthm}
\usepackage{mathtools}
\usepackage{txfonts}
\usepackage{mathdots}
\usepackage{bbm}

\usepackage{graphicx}
\usepackage{subcaption}
\usepackage{booktabs}
\usepackage{longtable}
\usepackage{tabularx}
\usepackage{multirow}
\usepackage{makecell}
\usepackage{pdflscape}
\usepackage{rotating}
\usepackage{float}
\usepackage{placeins}

\usepackage{algpseudocode}

\usepackage[nottoc]{tocbibind}

\usepackage{authblk}

\usepackage{hyperref}
\hypersetup{
    colorlinks=true,
    linkcolor=blue,
    citecolor=blue,
    urlcolor=blue
}

\usepackage{enumitem}

\usepackage{url}
\usepackage{comment}
\usepackage{setspace}
\usepackage{geometry}
\usepackage{varwidth}
\usepackage{footnote}
\usepackage[normalem]{ulem}

\makesavenoteenv{table}

\newcolumntype{?}{!{\vrule width 1.2pt}}

\def\BibTeX{{\rm B\kern-.05em{\sc i\kern-.025em b}\kern-.08em
    T\kern-.1667em\lower.7ex\hbox{E}\kern-.125emX}}

\newcommand{\added}[1]{\textcolor{black}{#1}}
\newcommand{\modified}[1]{\textcolor{black}{#1}}
\newcommand{\secondround}[1]{\textcolor{black}{#1}}

\makeatletter
\renewcommand\AB@affilsepx{, \protect\Affilfont}
\makeatother

\usepackage{tabularx}

\providecommand{\keywords}[1]{%
  \small%
  \textbf{\textit{Keywords---}} #1%
}

\newcommand{\xvect}{\mbox{\bf x}}

\newcommand{\unvect}{\mbox{\bf 1}}

\newcommand{\Avect}{\mbox{\bf A}}

\newcommand{\Fvect}{\mbox{\bf F}}
\newcommand{\Gvect}{\mbox{\bf G}}
\newcommand{\Ivect}{\mbox{\bf I}}

\newcommand{\Kvect}{\mbox{\bf K}}
\newcommand{\Lvect}{\mbox{\bf L}}
\newcommand{\Hvect}{\mbox{\bf H}}
\newcommand{\Pvect}{\mbox{\bf P}}

\newcommand{\Wvect}{\mbox{\bf W}}
\newcommand{\Uvect}{\mbox{\bf U}}

\newcommand{\Mvect}{\mbox{\bf M}}
\newcommand{\Svect}{\mbox{\bf S}}
\newcommand{\Yvect}{\mbox{\bf Y}}
\newcommand{\Xvect}{\mbox{\bf X}}
\newcommand{\Zvect}{\mbox{\bf Z}}

\begin{document}
\title{\textbf{Advanced Unsupervised Learning: A Comprehensive Overview of Multi-View Clustering Techniques}}
\author[1]{Abdelmalik Moujahid}
\author[2, 3]{Fadi Dornaika\thanks{Corresponding author}}

\affil[1]{\textit{Universidad Internacional de La Rioja (UNIR)}}
\affil[2]{\textit{University of the Basque Country}}
\affil[3]{\textit{IKERBASQUE, Basque Foundation for Science}}

\affil[ ]{

\small \texttt{abdelmalik.moujahid@unir.net, fadi.dornaika@ehu.eus}}

\date{}
\maketitle
\begin{abstract}
Machine learning techniques face numerous challenges to achieve optimal performance. These include computational constraints, the limitations of single-view learning algorithms and the complexity of processing large datasets from different domains, sources or views. In this context, multi-view clustering (MVC), a class of unsupervised multi-view learning, emerges as a powerful approach to overcome these challenges. MVC compensates for the shortcomings of single-view methods and provides a richer data representation and effective solutions for a variety of unsupervised learning tasks. In contrast to traditional single-view approaches, the semantically rich nature of multi-view data increases its practical utility despite its inherent complexity. This survey makes a threefold contribution: (1) a systematic categorization of multi-view clustering methods into well-defined groups, including co-training, co-regularization, subspace, deep learning, kernel-based, anchor-based, and graph-based strategies; (2) an in-depth analysis of their respective strengths, weaknesses, and practical challenges, such as scalability and incomplete data; and (3) a forward-looking discussion of emerging trends, interdisciplinary applications, and future directions in MVC research. This study represents an extensive workload, encompassing the review of over 140 foundational and recent publications, the development of comparative insights on integration strategies such as early fusion, late fusion, and joint learning, and the structured investigation of practical use cases in the areas of healthcare, multimedia, and social network analysis. By integrating these efforts, this work aims to fill existing gaps in MVC research and provide actionable insights for the advancement of the field.
\end{abstract}

 \hspace{10pt}

\section{Introduction}

Multi-view learning focuses on using information from different sets of features or representations, called {\em views}, to improve learning performance. The basic idea is that different views of the same data provide unique and complementary insights, so it is beneficial to consider them simultaneously. The field addresses challenges such as dealing with missing or noisy views, ensuring alignment between views, and selecting appropriate integration techniques.

In multi-view clustering, the way in which the information from multiple views is integrated has a significant impact on the clustering results. Depending on when and how the views are combined, the methods for multi-view clustering can be broadly categorized into three primary integration strategies: \textbf{early fusion}, \textbf{late fusion} and \textbf{joint learning}. \secondround{The three categories serve as an overarching framework for understanding multi-view clustering (MVC) approaches, offering a conceptual foundation from which various methodologies can be explored. While these categories provide a general overview, the subsequent sections of this paper delve into specific techniques and advancements within each category, examining how recent innovations and refinements have addressed the limitations and enhanced the performance of multi-view clustering methods across a variety of applications. This structure aims to provide readers with both a broad understanding and a detailed, in-depth perspective on the evolving landscape of multi-view clustering. The three primary categories are as follows:}

\begin{itemize}
 \item \textbf{Early Fusion}:
 \added{Early fusion methods combine the features from multiple views at the input level by merging the feature representations from all views into a single, unified feature model \cite{Liang2022,Liang2019}. This approach allows the model to learn from all available data simultaneously, processing the combined information early in the learning process.} 
 
 \added{One of the main advantages of early fusion is its computational efficiency, as it allows a single model to work with the combined feature set. However, early fusion assumes that all views are perfectly aligned and equally informative, which is often not the case in practice. In real-world applications, the views may be noisy or contain incomplete information, which can negatively impact the performance of early fusion methods. In addition, this approach cannot handle heterogeneity between views (e.g. different data modalities) as effectively as other strategies.}
 \item \textbf{Late Fusion}:
 \added{Late fusion approaches take a different strategy by training separate models for each view independently and combining their results — such as predictions or cluster assignments — at a later stage. In this way, each view can be processed individually, providing flexibility in dealing with heterogeneous data types or situations where some views might be missing \cite{Li-Miaomiao2024,Bruno2009736}. }
 
 \added{For example, in a clustering problem with multiple views, the individual clustering results from each view could be merged into a final consensus clustering. While late fusion is highly adaptive and robust to incomplete or varying data types, it cannot capture complex dependencies and interactions between the different views. Since each model is trained independently, the relationships between views are not always well utilized, which can lead to suboptimal integration of the data.}
 \item \textbf{Joint Learning}:
 \added{Joint learning techniques aim to simultaneously learn a common representation or task-specific features for all views, often by mapping each view into a common latent space. These methods explicitly capture the interactions and complementarities between views during the learning process, promoting an integrated understanding of the data \cite{Li2022,Huang-Aiping2021}.}
 
 \added{Joint learning approaches can be particularly powerful when the views are complementary, as they allow the model to exploit the full potential of the available information. By learning a common representation or model, these techniques improve the matching and integration of data from different views, which can lead to better performance than early or late fusion methods. However, joint learning is often more computationally intensive and requires more sophisticated optimization techniques and larger computational resources. It can also depend on the choice of model architecture, and the learning process can be more complex and prone to overfitting, especially when the data is noisy or sparse.}
\end{itemize}

\added{Each of these strategies has its own strengths and limitations, making them suitable for different application scenarios. For example, early fusion methods are effective when the views are well aligned and have similar characteristics, such as in multi-channel image processing. Late fusion is often used in scenarios with heterogeneous data types, such as in bioinformatics, where each view represents a different biological modality. Joint learning is particularly useful in applications such as sensor networks or multitask learning, where capturing interactions between different views is crucial.}

\added{This review systematically examines these strategies and provides a comparative analysis of the underlying principles, strengths and limitations of early fusion, late fusion and joint learning methods in multi-view clustering. The goal is to provide researchers with a comprehensive understanding of the multi-view clustering landscape and help them identify the most appropriate integration strategy for specific applications.}

The emerging field of multi-view clustering techniques specifically aims to seamlessly integrate this heterogeneous data landscape and extract meaningful patterns from it. By considering multiple data perspectives, these techniques have the potential to outperform traditional single-view clustering approaches and lead to more accurate and robust cluster mappings, as highlighted in studies by Xu et al. (2015) and Kumar et al. (2011) \cite{Xu2015, Kumar2011}. Furthermore, the application of multi-view clustering goes beyond improved clustering accuracy. It enables a deeper understanding of complex data by revealing hidden structures, relationships and insights that may remain undetected when analyzing single perspectives in isolation, as Cai et al. \cite{Cai2011} emphasize. This synergy between multi-view learning and clustering techniques opens up avenues for improved data analysis and knowledge extraction in various real-world applications.

\added{
\subsection{Addressing the Importance of Multi-view Clustering}
Multi-view clustering has become a central approach in the field of unsupervised learning, driven by the increasing prevalence of heterogeneous and multimodal datasets in modern applications. In contrast to single-view learning methods that rely on a single representation or perspective of the data, multi-view clustering utilizes multiple complementary views to provide a more holistic understanding of the underlying structures. This paradigm is particularly important for several reasons:
\begin{enumerate}
 \item \textbf{Overcoming limitations in data representation:}
 Single-view methods often fail to capture the variety and richness of data coming from different sources or modalities. Multi-view clustering addresses this limitation by integrating complementary information from multiple views, resulting in a more robust and comprehensive data representation. In biomedical research, for example, combining genomic data (one view) with proteomic data (another view) leads to a more comprehensive understanding of biological processes and improves the clustering of disease subtypes.
 \item \textbf{Handling heterogeneous datasets:}
 In real-world scenarios, datasets often comprise different domains, each of which provides unique insights. In multimedia analysis, for example, a dataset may contain text descriptions, audio recordings and visual features of videos. Multi-view clustering processes these heterogeneous modalities to effectively group similar multimedia content, outperforming single-view approaches that only consider one modality at a time.
 \item \textbf{Reducing bias through single-view dependency:}
 Dependence on a single view can introduce bias and limit generalizability, as the selected view may not fully represent the variability of the data. Multi-view clustering synthesizes information across multiple views to mitigate this problem. For example, when analyzing social networks, a single-view approach may only consider user interactions (e.g. likes or shares), while multi-view clustering integrates other perspectives such as user profiles and content preferences, resulting in more accurate community detection.
\end{enumerate}
By systematically addressing these constraints, multi-view clustering provides a transformative framework that meets the needs of modern machine learning tasks. Its ability to fuse complementary information, align different data modalities, and improve clustering performance highlights its crucial role in advancing unsupervised learning methods.
}

\subsection{Challenges}
\added{Multi-view clustering addresses the challenge of heterogeneity in real-world datasets by integrating different types of information and perspectives into a unified framework \cite{Xu2015}. By bridging the semantic gap between simple and high-level features, it enables the discovery of meaningful patterns, especially in multimedia data \cite{Cai2011}. However, despite its advantages, multi-view clustering faces several interrelated challenges that need to be overcome for effective application:}

\added{One key challenge in clustering with multiple views is the high-dimensional feature space that results from the integration of multiple views. While additional information improves clustering, it also introduces the "curse of dimensionality", which can degrade performance and increase computational complexity. Dimensionality reduction techniques, such as principal component analysis (PCA) \cite{greenacre2022principal} and t-SNE \cite{maaten2008visualizing}, reduce the dimensionality of individual views. Advanced methods such as multi-view subspace learning \cite{xu2013survey} uncover common latent representations across all views and thus improve the efficiency of clustering.}

\added{To efficiently process large datasets, anchor-based methods provide a scalable approach by selecting a small set of representative data points, called anchors, to approximate the original dataset. These methods effectively reduce computational complexity while preserving the structural relationships within the data. Anchor Graph Regularization (AGR) \cite{Yang2022} uses these anchors to construct similarity graphs that enable efficient computations and preserve the integrity of the data structure. In the context of multi-view clustering, anchor-based techniques extend this principle by identifying anchors that capture common structures across multiple views \cite{Ou-Qiyuan2020,Zhou2025,Li10345742,Wang-Siwei2022}, which promotes the discovery of common patterns and ensures computational efficiency.}

\added{Apart from dimensionality, the diversity and heterogeneity of multi-view datasets require more sophisticated methods to uncover complex relationships. Tensor factorization models have proven useful for clustering and merging data across modalities \cite{Chen2019,Chen2021,xia2022tensor}. Techniques such as Tucker decomposition \cite{tucker1966some} and Kolda and Bader's framework \cite{kolda2009tensor} reduce dimensionality while identifying common latent structures. Sequentially Truncated Higher-Order Singular Value Decomposition (ST-HOSVD) \cite{fang2019sequentially} provides robust data reconstruction and clustering, while hybrid methods combining Dynamic Mode Decomposition (DMD) with Graph Laplacian techniques \cite{zhu2022dynamic} are promising for large-scale multi-view clustering. These tensor-based approaches complement anchor-based methods and provide powerful solutions to the challenges posed by high-dimensional, heterogeneous data.}

\added{In addition to the challenges associated with dimensionality and heterogeneity, several other issues need to be addressed for the effective application of multi-view clustering}

\begin{itemize}
 \item \added{\textbf{Model selection and scalability:}}
 \added{Choosing an appropriate multi-view learning model and fusion strategy is a non-trivial task as it depends on the specific problem and the dataset. In addition, some multi-view learning algorithms have difficulty scaling effectively to large datasets, which limits their practical utility \cite{liu2013sleec}.}

 \item \added{\textbf{Merging clustering results:}}
 \added{A key challenge is to merge the clustering results from different views in a natural and coherent way. Since each view provides a different perspective on the data, integrating these results requires novel clustering target functions that can capture the diverse information from multiple sources.}

 \item \added{\textbf{Determining the importance of views:}}
 \added{Not all views contribute equally to the clustering process. One of the challenges of clustering with multiple views is determining the relative importance of the individual views. This requires mechanisms that weight the views appropriately and ensure that the more informative views have a greater influence on the clustering result}

 \item \added{\textbf{Heterogeneity of data with multiple views:}}
 \added{In real-world applications, multi-view data often exhibits significant heterogeneity in terms of scale, modality and quality. This heterogeneity poses a major challenge in multi-view learning \cite{Yan2016}. In particular, evaluating the correlation and redundancy between the different views is crucial, as overly correlated views can lead to suboptimal clustering results \cite{cai2013heterogeneous}}

 \item \added{\textbf{Incomplete Multi-view Clustering (IMC):}}
 \added{Addressing the challenge of missing data has led to the emergence of Incomplete Multi-view Clustering (IMC). However, most existing multi-view clustering methods assume complete data, which makes them less effective in real-world scenarios where data is often incomplete. Developing methods that can deal with missing data, especially with a high rate of missing views, remains an ongoing challenge \cite{liu2020efficient, lv2022viewconsistency}.}
\end{itemize}

\setlength{\columnsep}{15pt}  
\renewcommand{\arraystretch}{1}  

\begin{table*}[ht]
\added{
\centering
\scriptsize  
\resizebox{\textwidth}{!}{
\begin{tabular}{|p{1cm}|p{1cm}|p{3cm}|p{5cm}|p{6cm}|}  
\hline
\textbf{Year} & \textbf{Ref.}          & \textbf{Title}                                     & \textbf{Main Focus}                                                                                                                            & \textbf{Key Contributions}   \\ \hline
2025 & \cite{Qin2025}             & A Survey on Representation Learning for Multi-view Data  & Focuses on multi-view clustering, self-supervised multi-view clustering and incomplete multi-view clustering.  & Provides a novel survey on multi-view clustering by organizing existing algorithms into two distinct categories: non-self-supervised and self-supervised multi-view clustering, addressing the gap left by previous surveys that overlooked simultaneous consideration of both. \\ \hline
2024 & \cite{HARIS2024107857}     & Breaking down multi-view clustering: A comprehensive review of multi-view approaches for complex data structures  & Provides a comprehensive classification of Multi-View Clustering (MVC) methods, categorizing them into generative and discriminative approaches, with a focus on deep learning-based techniques.                                                                                         & Classifies MVC methods into generative and discriminative categories, emphasizing deep learning's role in complex data structures. Provides a systematic comparison and identifies research gaps. \\ \hline
2024 & \cite{RAYA2024128348}      & Multi-modal data clustering using deep learning: A systematic review                                            & Focuses on introducing a novel taxonomy for deep learning-based multi-modal clustering.                                            & Explores CNNs, Autoencoders, RNNs, and GCNs, identifies gaps in multi-modal clustering research, and suggests future research directions. \\ \hline
2023 & \cite{Fang2023}      & A Comprehensive Survey on Multi-View Clustering  & The survey categorizes current MVC approaches into two main technical Mechanisms: heuristic-based multi-view clustering (HMVC) and neural network-based multi-view clustering (NNMVC).                                            & Explores key approaches within HMVC, including nonnegative matrix factorization, graph learning, and tensor learning, as well as deep representation learning and deep graph learning in NNMVC. \\ \hline
2022 & \cite{Wen2022SurveyIM}      & A Survey on Incomplete Multiview Clustering   & Focuses on incomplete multi-view clustering (IMC). IMC is particularly relevant for practical applications such as disease diagnosis, multimedia analysis, and recommendation systems, where incomplete data is common & The survey unifies IMC methods under common frameworks, conducts a comparative analysis of representative approaches, and highlights open problems to guide future research. \\ \hline
2021 & \cite{De-Handschutter2021} & A survey on deep matrix factorizations             & Focuses on deep matrix factorization techniques and their applications.                                                                             & Explores advanced matrix factorization methods in deep learning, highlights applications in multidimensional data analysis, and identifies future challenges.      \\ \hline
2021 & \cite{Chao2021}           & A Survey on Multiview Clustering                   & Review of multiview clustering methods, proposing a novel taxonomy based on generative and discriminative approaches.                               & Classifies multiview clustering approaches into generative and discriminative categories; links MVC with representation, ensemble clustering, multi-task learning. \\ \hline
2020 & \cite{FU2020148}            & An overview of recent multi-view clustering        & Review of recent multiview clustering algorithms with an experimental focus.                                                                        & Divides algorithms into three main categories; conducts extensive experiments on seven datasets using accuracy, NMI, and purity metrics; proposes future directions.  \\ \hline
2018 & \cite{Yang2018}         & Multi-view Clustering: A Survey                    & Review of multiview clustering algorithms, organized according to mechanisms and underlying principles.                                              & Proposes a taxonomy in five categories: co-training, multi-kernel learning, graph-based clustering, subspace clustering, and multitask multiview clustering.        \\ \hline
\end{tabular}
}
\caption{\added{Summary of existing surveys on multi-view clustering methods, highlighting their focus on specific subfields, applications, or methodologies. The table underscores the need for a recent and comprehensive survey, like ours, to address gaps in interdisciplinary approaches and provide a unified perspective on the evolving landscape of multi-view clustering techniques and applications.
\label{tab:multi_view_clustering_surveys}}}}
\end{table*}
\added{Table \ref{tab:multi_view_clustering_surveys} provides an overview of key surveys on multi-view clustering (MVC), summarizing their main focus and contributions. Existing surveys often focus on specific subfields, applications, or disciplinary boundaries, leading to gaps in coverage and overlooking new interdisciplinary approaches. Each of these surveys presents a distinct perspective on MVC techniques, ranging from deep matrix factorizations and generative versus discriminative approaches to the categorization of algorithms and their evaluation in real-world applications. The surveys address various challenges in MVC, such as scalability and noise management, and propose future research directions to advance the field. Consequently, there is a compelling need for a systematic and up-to-date survey that bridges these gaps, offering a comprehensive overview of multi-view clustering methods and their applications across diverse domains. This table highlights the breadth of approaches and provides a comparative look at the state-of-the-art in multi-view clustering.}

\subsection{Motivation}
The motivation for conducting this survey lies in the recognition of multi-view learning as a central and effective approach to overcoming the limitations of single-view methods. Multi-view learning not only compensates for these limitations, but also provides a more comprehensive representation of the data and thus solutions to various machine learning challenges. The versatility of multi-view learning makes it an invaluable tool that can be used in various domains and applications.

In contrast to traditional data representations that represent objects from a single view, multi-view data is semantically rich, which increases its practical utility despite its higher inherent complexity. In light of these considerations, this survey aims to provide a systematic overview of the main clustering methods documented in the scientific literature for processing multi-view data. Moreover, it contributes to a better understanding of the landscape of multi-view learning methods, their applications and their potential importance in addressing the challenges faced by today's machine learning techniques.

On the other hand, the lack of comprehensive survey papers in the literature addressing multi-view clustering arises from the diverse methodological landscape, rapid evolution of techniques, application-specific focus, and the overall growth of the field. Existing surveys \cite{De-Handschutter2021,Chao2021,Yang2018,FU2020148} often focus on specific subfields, applications or disciplinary boundaries, leading to gaps in coverage and overlooking new interdisciplinary approaches. Consequently, there is a compelling need for a systematic and up-to-date survey that bridges these gaps, offering a comprehensive overview of multi-view clustering methods and their applications across diverse domains.

\added{
\subsection{Structure of the Survey}
This review is systematically organized to provide a comprehensive overview of the multi-view clustering landscape, structured as follows:
\begin{enumerate}
    \item \textbf{Introduction}  
    The opening section highlights the growing importance of multi-view clustering in modern machine learning and data analytics. It discusses the significance, challenges, and motivations driving this area of research.    
    \item \textbf{Fundamentals of Multi-view Clustering} (Section \ref{Fundamentals})  
    This section lays the groundwork by explaining the key principles of multi-view clustering. It presents a taxonomy of multi-view approaches to help contextualize the methods discussed later.   
    \item \textbf{Classical Methods for Multi-view Clustering} (Section \ref{Classical-methods})  
    We categorize and explain seminal approaches in multi-view clustering, focusing on classical techniques that have shaped the field.      
    \item \textbf{Exploring Graph-based Multi-view Clustering} (Section \ref{summary-table})  
    We provide a comprehensive overview of novel graph-based algorithms tailored to the challenges of multi-view clustering, such as handling missing data, noise reduction, and efficient computation.    
    \item \textbf{Multi-view Clustering with Missing or Incomplete Data} (Section \ref{mvc-missing-data})  
    This section addresses the recent challenges and methods for clustering with incomplete data, a common issue in multi-view settings.    
    \item \textbf{Formal Review of Typical Approaches} (Section \ref{ReviewTypicalApproach})  
    A rigorous mathematical examination of the typical multi-view clustering approaches, focusing on the formal framework and methods used.    
    \item \textbf{Datasets for Multi-view Clustering} (Section \ref{tab:dataset-info})  
    We provide an overview of key datasets used to evaluate multi-view clustering methods, detailing common feature extraction techniques and considerations for selecting the optimal number of views.    
    \item \textbf{Conclusion}  
    This section summarizes the main findings of the review and reinforces the importance of multi-view clustering in modern machine learning.
    \item \textbf{Future Directions} (Section \ref{future-directions})  
    The final section discusses the potential future developments and research opportunities in multi-view clustering.
\end{enumerate}
}

\section{Fundamentals of Multi-view Clustering}
\label{Fundamentals}

\modified{Multi-view learning has gained much attention due to its potential to improve the performance of models by leveraging multiple data perspectives. Integrating information from different views increases prediction accuracy and provides a more comprehensive understanding of the data structure, improving the robustness of the clustering process \cite{xu2013survey}. By capturing the underlying structures from different views, multi-view methods can often perform better than single-view approaches \cite{Kumar2011}. Furthermore, learning from multiple views is crucial for domain adaptation, i.e. the alignment of data representations from different sources or domains \cite{Sun2013}.}

\modified{Despite its advantages, multi-view clustering faces several challenges. The heterogeneity of data in terms of scale, modality and quality can make it difficult to effectively merge different views. In addition, overly correlated views can affect the performance of clustering algorithms and lead to suboptimal results. High-dimensional feature spaces resulting from the integration of multiple views require the use of dimensionality reduction techniques to avoid problems such as the "curse of dimensionality" \cite{Sindhwani2008}. In addition, model selection and scalability remain a challenge, as the choice of an appropriate multi-view learning model depends on the specific problem and the dataset, and some models do not scale well to large datasets \cite{Liu2013}.}

In the field of multi-view clustering, established methods can be systematically categorized into different groups, each tailored to a specific strategy for merging information from different views. We emphasize that the boundaries between the categories may be blurred, as a given clustering approach may belong to more than one category.

\subsection{Taxonomy of Multi-View Clustering Approaches}

Given the extensive diversity of multi-view clustering methods and their varying approaches to view integration, our systematic review organizes this methodological landscape into a comprehensive taxonomy, presented in Table~\ref{tab:full-taxonomy-landscape}.

 Below, we categorize these methods into distinct groups based on their underlying principles and techniques. While these categories provide a structured overview, it is important to note that certain approaches may overlap multiple categories.

\begin{itemize}
    \item \textbf{Co-training}: This method starts by clustering data from a single view and iteratively refines the results across other views. It is advantageous because it allows for the enhancement of initial clustering through additional views. However, it requires a strong initial clustering and may struggle when views are highly inconsistent or noisy \cite{Kumar2011, Tan2020}.
    
    \item \textbf{Co-regularized multi-view spectral clustering}: This technique integrates multi-view learning with spectral clustering by introducing a co-regularization term, harmonizing clusterings across different views. The benefit is that it can effectively combine the views' information in an unsupervised manner. However, the choice of the appropriate regularization term can be challenging and may affect the clustering performance significantly \cite{Kumar2011}.

    \item \textbf{Kernel-based multi-view clustering}: Kernel methods map data into high-dimensional spaces to handle non-linearity, making it easier to cluster data that are not linearly separable. This method is useful for addressing the diversity in shapes across views. However, learning optimal kernels and performing dimensionality reduction are computationally expensive tasks, and the risk of overfitting increases as the number of views grows \cite{Li2016, Chen2019, Wang2021}.

    \item \textbf{Subspace multi-view clustering:}
    Subspace multi-view clustering aims to learn a unified feature representation by assuming that all views share this representation. It can be divided into subspace-based methods \cite{Wang2015, Chen2020} and matrix factorization approaches \cite{Khan2022}, both of which are designed to analyze low-dimensional representations embedded in multiple views. \added{However, conventional methods often struggle to capture high-dimensional information from nonlinear subspaces or overlook the high-level relationships between fundamental partitions obtained by clustering in a single view. These relationships could bridge the gap between heterogeneous feature spaces and improve clustering performance.}
    
    \added{Moreover, existing multi-view ensemble clustering methods often overlook the noise that arises in the data generation phase. To overcome this challenge, recent approaches have introduced new methods that incorporate regularization techniques and denoising strategies. For example, Zheng et al. \cite{ZHENG2024121187} proposed a multiview subspace clustering method that combines hypergraph p-Laplacian regularization with low-rank subspace learning. This method was developed to capture complex hierarchical structures in the data while effectively mitigating the effects of noise. }
    
    \added{By integrating ensemble strategies, noise reduction and weight adaptation, these techniques improve the robustness of multi-view clustering and lead to better performance in heterogeneous and noisy environments. Similar approaches that fuse regularization, denoising, and low-rank learning in multi-view environments can be found in recent studies such as \cite{Pu-Xinyu2023}, which propose a robust low-rank graph multi-view clustering method that integrates spectral embedding, non-convex low-rank approximation and noise handling.}

    \item {\bf Deep learning based approach}: The integration of deep learning with multi-view clustering leverages neural networks to create joint representations, significantly improving clustering performance in datasets with multiple views. \added{However, current models have been criticized as shallow, as they directly map multi-view data to low-dimensional space, often neglecting essential nonlinear structure information within each view \cite{Liu2021}. }
    
    \added{To approximate these limitations, the authors in \cite{GUPTA2024109597} propose DeConFCluster, an unsupervised multi-view clustering framework based on Deep Convolutional Transform Learning (CTL). By eliminating the need for an additional decoder network during training, DeConFCluster reduces overfitting in data-constrained scenarios, a common drawback of encoder-decoder-based methods. Furthermore, the model incorporates a K-Means-inspired loss function, enhancing representation learning for clustering tasks. The framework outperforms state-of-the-art multi-view deep clustering techniques on five benchmark datasets, showcasing its efficacy in capturing both joint representations and nonlinear structural information. }

    \added{\item \textbf{Graph-based multi-view clustering:}
    Graph-based methods integrate multiple views by constructing a unified graph from individual similarity matrices. The relationship between the data points of each view is represented as a graph, and the fusion of these graphs helps to capture complex interactions between the views \cite{Wang2018, Tang2018}. This approach is very powerful for managing diverse relationships, but it also brings computational challenges, especially in graph construction and matrix fusion, especially for large datasets \cite{Zhang2021}.}
    
    \added{A key problem is the selection of relevant views for graph creation, as not all views contribute equally to clustering. Some views may cause noise or redundancy, so it is important to identify and prioritize the most informative views. Methods such as adaptive graph learning and regularization techniques have been proposed to address these issues and optimize both computational efficiency and clustering accuracy \cite{Wang2018, Zhang2021}.}

     \added{\item \textbf{Anchor-based methods:}
     meet the main objectives of multi-view clustering as they provide a scalable approach to processing large data sets. By selecting representative samples (anchors) to approximate the entire dataset, scalability challenges are effectively addressed while preserving essential data structures. This approach improves scalability by reducing computational complexity and promotes efficiency when processing diverse, large-scale data sources. Anchors also contribute to robustness by capturing important patterns in different views, ensuring consistent performance even in the presence of noise or incomplete data. In addition, they enable flexibility as they can be seamlessly adapted to different clustering frameworks and methods. Ultimately, anchor-based methods promote effectiveness by preserving the quality of clustering results while optimizing resource utilization.}  

    \added{A notable technique in this area is Anchor Graph Regularization (AGR) \cite{Yang2022}, which constructs a similarity graph using the selected anchors and enables efficient computations. In multi-view clustering, these methods are extended by selecting anchors that encapsulate common information across multiple views, which improves computational efficiency and data representation. Recent research, including work by \cite{Ou-Qiyuan2020,Zhou2025,Li10345742,Wang-Siwei2022}, has further explored anchor-based approaches, highlighting their ability to identify common patterns across different views and provide practical solutions for large-scale clustering tasks.}

    \added{\item {\bf Other approaches} address the challenges of multi-view clustering by tackling specific limitations in traditional methods. For instance, the work \cite{XU2024111590} focuses on two significant issues: (1) the inability of hard clustering techniques to capture uncertainty between samples and clusters, and (2) the challenge of effective incremental learning when the number of views increases. To address these, the study introduces a three-way fuzzy spectral clustering algorithm that generates soft clustering results, effectively modeling uncertainty. Furthermore, it incorporates an incremental learning mechanism based on sequential decision-making to handle dynamically increasing views. By combining these advancements, the proposed multi-view clustering algorithm based on sequential three-way decision-making achieves enhanced clustering accuracy and efficiency, as validated through experimental evaluations.}   
\end{itemize}

\added{\subsection{Practical Applications of Multi-view Clustering}}
\label{sec:practical-applications}

\added{Multi-view clustering has found wide application in a number of areas where the integration of multiple data views enables a more comprehensive understanding of the problem at hand. This section highlights the main areas where multi-view clustering methods have been used effectively:}

\added{ \begin{itemize}
 \item \textbf{Healthcare and medical diagnostics:}
Multiview clustering methods have proven their effectiveness on various types of data, including medical imaging, multi-omics data and physiological signals such as EEG. In medical imaging, different imaging modalities (e.g. MRI, CT scans, PET scans) are combined to improve diagnostic accuracy. Similarly, when integrating multi-omics data, consistent clustering patterns (common to all omics levels) can be identified alongside differential patterns (specific to individual omics types), helping to uncover biologically meaningful correlations and insights into disease mechanisms \cite{AlKuhali2022}. Multi-view clustering has also been successfully applied to EEG signals and enables the classification of brain activity into different patterns, such as seizures and non-seizures, thus improving diagnostic precision \cite{Zhan2020}. In addition, these techniques are used to identify disease subtypes, support personalized treatment planning and predict patient outcomes by integrating multiple clinical and biological data sources \cite{PFEIFER2023104406, Feng2022}. \cite{AlKuhali2022}.
 \item \textbf{Social network analysis:}
 In social networks, multi-view clustering can be used to analyze data from different sources such as social interactions, user behavior and content analysis. By combining multiple views, e.g. interactions between users, posts and metadata, multi-view clustering enables more accurate identification of communities, influencers or trends within the network \cite{Kim2017,Lan2017,Dong2013}.
 \item \textbf{Computer vision and image processing:}
 Multi-view clustering has been widely applied in computer vision, especially in object recognition and scene analysis. These tasks benefit from the ability to integrate information from multiple views or modalities to improve the accuracy of the models. One notable application is the tracking of pedestrians in crowded environments where targets are temporarily blocked and reappear. To solve this problem, a novel multiview clustering method has been developed that improves tracking accuracy by utilizing the correlation of fusion features within a multiview system. This method is particularly beneficial in cases where pedestrians are occluded or reappear after disappearing briefly \cite{Chen2024204}. In object detection and scene analysis, multi-view clustering enhances recognition accuracy by integrating multiple views of the same scene. Techniques employing multi-view learning, such as those using deep neural networks, effectively capture inter-view relationships, leading to improved performance in complex tasks like 3D object recognition and scene segmentation. These methods have proven successful in dynamic environments, offering significant advancements in scene understanding and recognition accuracy \cite{Teepe2024,Li-Jingwen2023}.
\item \textbf{Natural language processing:}  
    In natural language processing, multi-view clustering has been successfully applied to document clustering and sentiment analysis. These applications often involve integrating different views of text data, such as syntactic and semantic information, to improve the quality and relevance of clustering results. Multi-view clustering techniques have shown effectiveness in these tasks by capturing diverse features from the text, which helps in grouping documents with similar content or sentiment \cite{Zhou2007, Zhan2020, Serra2018265, Hsueh2020275, Yu2023}. 
 \item \textbf{Recommendation systems and personalized medicine:}
Multi-view clustering plays a crucial role in data fusion as it enables the integration of different data views for more informed decision making. In recommendation systems, multi-view clustering methods enable the combination of different user data sources (e.g. behavior, preferences and demographics) to provide personalized recommendations. Multi-view clustering has also been used in personalized medicine to identify disease subtypes by combining different medical views (e.g. imaging data and genetic information). For example, a multi-view learning approach has been developed to identify imaging-related subtypes in mild cognitive impairment (MCI). The approach uses techniques such as Deep Generalized Canonical Correlation Analysis (DGCCA) to learn low-dimensional correlated embeddings, which significantly advances personalized medicine and medical diagnostics \cite{Feng2022}.
\end{itemize}
The inclusion of these practical applications emphasizes the versatility and impact of multiview clustering techniques in practice and demonstrates their potential in various domains. This section provides readers with both an overview of existing approaches and practical insights into how these methods effectively address complex, real-world challenges.}

\secondround{In the following sections, we review different methodological approaches to multi-view clustering (MVC), each addressing distinct aspects of the problem rather than forming a step-by-step optimization process. Section 3 presents classical MVC methods that serve as the foundation for more advanced techniques. Section 4 focuses on graph-based approaches, which incorporate graph structures to enhance clustering performance by capturing relationships across views. Finally, Section 5 explores MVC techniques designed to handle missing or incomplete data, addressing a critical challenge in real-world applications. These three sections provide complementary perspectives on MVC, highlighting different methodological advancements tailored to specific challenges in multi-view learning.}

\newpage
\begin{landscape}
\begin{longtable}
{|p{0.6cm}|p{0.8cm}|p{3cm}|p{11cm}|p{5.5cm}|}
\caption{Compilation of Multi-View Clustering Methods Developed in the Last Five Years.}  \label{tab:full-taxonomy-landscape} \\
\hline

\textbf{Ref.} & \textbf{Year} & \textbf{Methods} &\textbf{Multi-view Clustering Strategy} & \textbf{Dataset} \\
\hline
\endfirsthead

\multicolumn{4}{c}%
{{\tablename\ \thetable{} -- Continued from previous page}} \\
\hline
\textbf{Ref.} & \textbf{Year} &\textbf{Methods} & \textbf{Multi-view Clustering Strategy} & \textbf{Dataset} \\
\hline
\endhead

\hline \multicolumn{4}{|r|}{{Continued on next page}} \\ \hline
\endfoot

\hline
\endlastfoot
\cite{Zhou2025} & 2025 & \begin{itemize}[noitemsep,topsep=0pt, left=0.1cm] \item[\checkmark] Graph learning \item[\checkmark] Anchor-based\item[\checkmark] Bipartite graph  \end{itemize} & \begin{itemize}[noitemsep,topsep=0pt, left=0.1cm] \item Integrate anchor graph learning and subspace graph construction into a unified optimization framework based on a bipartite graph. \item It enhances clustering by jointly optimizing the projection matrix, consensus anchor matrix, and similarity matrix, ensuring connectivity constraints to form clusters directly.\end{itemize} & \begin{itemize}[noitemsep,topsep=0pt, left=0.1cm] \item Image datasets \item Multi-view datasets  \end{itemize} \\ \hline
\cite{Chen2025} & 2025 & \begin{itemize}[noitemsep,topsep=0pt, left=0.1cm] \item[\checkmark] Hypergraph \item[\checkmark] Non-negative matrix factorization\item[\checkmark] Tensor Schatten $p$-norm  \end{itemize}  & \begin{itemize}[noitemsep,topsep=0pt, left=0.1cm] \item Reconstructe missing views using a hypergraph, capturing both local structures and higher-order relationships. \item It integrates representation learning and clustering into a one-step framework, avoiding suboptimal results from two-step approaches.\end{itemize} & \begin{itemize}[noitemsep,topsep=0pt, left=0.1cm] \item Incomplete multi-view datasets  \end{itemize} \\ \hline
\cite{Dou2025} & 2025 & \begin{itemize}[noitemsep,topsep=0pt, left=0.1cm] \item[\checkmark] Multi-level graphs \item[\checkmark] Deep non-negative matrix factorization  \end{itemize} & \begin{itemize}[noitemsep,topsep=0pt, left=0.1cm] \item addresses the challenge of balancing diversity and consistency across multiple views. \item It integrates feature learning, multi-level topology representation, and clustering into a unified framework. Specifically, it uses deep non-negative matrix factorization (DNMF) to learn multi-level (hierarchical) representations of objects.\end{itemize} & \begin{itemize}[noitemsep,topsep=0pt, left=0.1cm] \item Image datasets \item Benchmark datasets  \end{itemize}  \\ \hline
\cite{Zhou2024} & 2024 & \begin{itemize}[noitemsep,topsep=0pt, left=0.1cm] \item[\checkmark] Sparse graph learning  \end{itemize} & \begin{itemize}[noitemsep,topsep=0pt, left=0.1cm] \item Addresses the challenge of allocating contributions of different views by assigning view-specific weights instead of equal weights. \item The method directly obtains cluster indicators by applying low-rank constraints, eliminating the need for post-processing.\end{itemize} & \begin{itemize}[noitemsep,topsep=0pt, left=0.1cm] \item Caltech101-07, Caltech101 \item UCI-digit, Mfeat, STL10 \end{itemize} \\ \hline
\cite{Zhao2024} & 2024 & \begin{itemize}[noitemsep,topsep=0pt, left=0.1cm] \item[\checkmark] Graph learning \item[\checkmark] Bipartite graph \item[\checkmark] Dynamic adaptation   \end{itemize} & \begin{itemize}[noitemsep,topsep=0pt, left=0.1cm] \item A learnable graph filter that dynamically refines the original feature space, progressively filtering out noise and producing a smooth, clustering-friendly representation. \item A unified bipartite graph that combines multi-granular structural information from different views, capturing both distinct and shared features across views.\end{itemize} & \begin{itemize}[noitemsep,topsep=0pt, left=0.1cm] \item Multiple benchmark datasets \end{itemize} \\ \hline
\cite{Wang2024} & 2024 & \begin{itemize}[noitemsep,topsep=0pt, left=0.1cm] \item[\checkmark] Deep learning  \end{itemize} & \begin{itemize}[noitemsep,topsep=0pt, left=0.1cm] \item combines the flexibility of deep learning with the statistical benefits of data-driven and knowledge-driven feature selection, providing interpretable results. \item It learns nonlinear relationships in multi-view data by using deep neural networks to create low-dimensional, view-independent embeddings, while imposing a regularization penalty on the reconstructed data. The method uses the normalized Laplacian of a graph to model bilateral relationships between variables within each view, promoting the selection of related variables. \end{itemize} & \begin{itemize}[noitemsep,topsep=0pt, left=0.1cm]\item Holm Breast Cancer, Study, LGG Dataset (grade 2 and 3) \item Shear Transformed, MNIST Dataset \end{itemize} \\ \hline
\cite{Li2024} & 2024 & \begin{itemize}[noitemsep,topsep=0pt, left=0.1cm] \item[\checkmark] Graph-based \item[\checkmark] Subspace-based\item[\checkmark] Kernel-based  \end{itemize} &\begin{itemize}[noitemsep,topsep=0pt, left=0.1cm] \item Incomplete multi-view subspace clustering based on multiple kernel completion, low-redundant representation learning, and weighted tensor low-rank constraint. \item The unified objective function combines the intact view-specific subspaces and the hidden low-rank tensor.\end{itemize} & \begin{itemize}[noitemsep,topsep=0pt, left=0.1cm] \item BBC-Sport, MSRCv1, \item 100Leaves, NGs, \item3Sources, ORL. \end{itemize}\\
\hline 
\cite{Su2023} & 2023 &\begin{itemize}[noitemsep,topsep=0pt, left=0.1cm] \item[\checkmark] Graph-based \item[\checkmark] Subspace-based\item[\checkmark] Anchor-based  \end{itemize} &\begin{itemize}[noitemsep,topsep=0pt, left=0.1cm] \item Anchor-based Multi-View Subspace Clustering with Graph Learning. \item Integrate anchor learning and the construction of coefficient matrices through a unified optimization procedure that exploits the global and local structure of the samples and the learned anchors.\end{itemize} & \begin{itemize}[noitemsep,topsep=0pt, left=0.1cm] \item Caltech101-7, Caltech101-20, \item SUN-RGBD, Animal, AwA, \item NUSWIDEOBJ, YoutubeFace. \end{itemize}\\
\hline
\cite{Pan2023}\footnote{See Section 6.4 for the formal definition of the method.} & 2023 &\begin{itemize}[noitemsep,topsep=0pt, left=0.1cm] \item[\checkmark] Graph-based \item[\checkmark] Anchor-based  \end{itemize}& \begin{itemize}[noitemsep,topsep=0pt, left=0.1cm] \item High-order multi-view clustering based on graph filtering, intrinsic relationships up to infinity order, adaptive graph fusion, and anchors selected by high-order structure\end{itemize}.& \begin{itemize}[noitemsep,topsep=0pt, left=0.1cm] \item ACM, DBLP, IMDB, \item Amazon Photos/Computers. \end{itemize}  \\
\hline
\cite{Zhong-Guo2023} & 2023 &\begin{itemize}[noitemsep,topsep=0pt, left=0.1cm] \item[\checkmark] Graph-based \end{itemize}&  \begin{itemize}[noitemsep,topsep=0pt, left=0.1cm] \item Integrate adaptive weighting, Laplacian embedding (spectral embedding), consensus graph learning and discrete indicator matrix learning into a unified framework. \item Outputs clustering results directly without the need for post-processing.\end{itemize} & \begin{itemize}[noitemsep,topsep=0pt, left=0.1cm] \item Leaves100, COIL100, NGs,\item  BBCSport, HW, ORL, \item Mfeat, ALOI. \end{itemize}  \\
\hline
\cite{Dornaika2023} & 2023 &\begin{itemize}[noitemsep,topsep=0pt, left=0.1cm] \item[\checkmark] Graph-based \item[\checkmark] Kernel-based  \end{itemize}& \begin{itemize}[noitemsep,topsep=0pt, left=0.1cm] \item Unified single-phase multi-view clustering with consensus graph learning and spectral representation. Jointly generates similarity graphs of the views and their joint similarity matrix using a unified global objective function. \item It takes as input a kernelized representation of the features and directly returns the individual graphs, the joint graph, the joint spectral representation and the cluster assignments.\end{itemize} & \begin{itemize}[noitemsep,topsep=0pt, left=0.1cm] \item ORL, COIL20, BBCSport, \item MSRCv1, MNIST-25000. \end{itemize}  \\
\hline
\cite{Dornaika2023c} & 2023 &\begin{itemize}[noitemsep,topsep=0pt, left=0.1cm] \item[\checkmark] Graph-based \item[\checkmark] Kernel-based  \end{itemize}& \begin{itemize}[noitemsep,topsep=0pt, left=0.1cm] \item Multi-view clustering via kernelized graph and nonnegative embedding. \item It is based on a single global criterion that jointly provides the consistent similarity matrix for all views, the consistent spectral representation, the soft cluster assignments and the view weights.\end{itemize} & \begin{itemize}[noitemsep,topsep=0pt,left=0.1cm] \item COIL20, ORL, Out-Scene, \item MNIST, BBCSport, MSRCv1, \item Caltech101-7, Extended-Yale. \end{itemize}  \\
\hline
\cite{Chen2023} & 2023 &\begin{itemize}[noitemsep,topsep=0pt, left=0.1cm] \item[\checkmark] Low-rank tensors  \end{itemize}& \begin{itemize}[noitemsep,topsep=0pt, left=0.1cm] \item An approach for learning low-rank tensors that can provide a consensus low-dimensional embedding matrix for incomplete multiview clustering.  \item It involves learning individual low-dimensional embedding matrices from incomplete multiview data, utilizing the self-expression property of high-dimensional data. \end{itemize}& \begin{itemize}[noitemsep,topsep=0pt, left=0cm] \item Reuters, O-Scene, \item Handwritten, COIL-20 \item ProteinFold, Flower17, \item SUN-RGBD, \item 100leaves, Caltech101 \end{itemize} \\\hline
\cite{Luo2022} & 2022 & \begin{itemize}[noitemsep,topsep=0pt, left=0.1cm] \item[\checkmark] Subspace Dual Clustering  \end{itemize} & \begin{itemize}[noitemsep,topsep=0pt, left=0.1cm] \item Combines dual-clustering and multiview subspace learning to simultaneously discover consensus representation and dual-clustering structure using alternating optimization.  \item A unified framework is developed to jointly explore clustering and subspace learning. \end{itemize} & \begin{itemize}[noitemsep,topsep=0pt, left=0cm] \item Real-world multiview dual \item single-clustering datasets \end{itemize}\\ \hline
\cite{Liu2022} & 2022 & \begin{itemize}[noitemsep,topsep=0pt, left=0.1cm] \item[\checkmark] Virtual-label Guided Matrix Factorization (VLMF)   \end{itemize} & \begin{itemize}[noitemsep,topsep=0pt, left=0.1cm] \item Utilizes graph regularization to capture geometric structure, and a virtual-label guided matrix factorization to recover and learn consensus latent representations.  \item The approach integrates clustering and latent representation learning into a joint optimization process. \end{itemize} & \begin{itemize}[noitemsep,topsep=0pt, left=0cm] \item Incomplete multi-view datasets \end{itemize}  \\ \hline
\cite{Wang2022} & 2022 & \begin{itemize}[noitemsep,topsep=0pt, left=0.1cm] \item[\checkmark]  Double embedding-transfer-based Multi-view Spectral Clustering (DETMSC)   \end{itemize} & \begin{itemize}[noitemsep,topsep=0pt, left=0.1cm] \item Incorporates two types of embeddings: consistency embedding and feature embedding.  \item Knowledge transfer between these embeddings is achieved via bipartite graph co-clustering, which improves clustering accuracy by learning both consistency across views and diversity of features.\item Robustness to noisy data is enhanced through sparse constraints.  \end{itemize} & \begin{itemize}[noitemsep,topsep=0pt, left=0cm] \item Real-world benchmark datasets \end{itemize} \\ \hline
\cite{Yin2022} & 2022 & \begin{itemize}[noitemsep,topsep=0pt, left=0.1cm] \item[\checkmark]  Anchor-based Incomplete Multi-view Spectral Clustering (AIMSC) \end{itemize}  & \begin{itemize}[noitemsep,topsep=0pt, left=0.1cm] \item Uses anchor points to connect instances from each view and recover missing data.  \item The similarities between data points are derived from their relationship with anchor points.\item The method then applies anchor-based spectral clustering to generate accurate clustering results.  \end{itemize} & \begin{itemize}[noitemsep,topsep=0pt, left=0cm] \item Multiple benchmark datasets \end{itemize}  \\ \hline
\cite{Chen-Man2022}\footnote{See Section 6.7 for the formal definition of the method.} & 2022 &\begin{itemize}[noitemsep,topsep=0pt, left=0.1cm] \item[\checkmark] Graph-based \item[\checkmark] Anchor-based  \end{itemize}& \begin{itemize}[noitemsep,topsep=0pt, left=0.1cm] \item A unified framework for anchor learning, graph construction and partitioning, while keeping the complexity almost linear.  \item Through mutual improvement, the model achieves a more discriminative and flexible anchor representation and cluster indicator. \end{itemize}& \begin{itemize}[noitemsep,topsep=0pt, left=0cm] \item Notting-Hill, \item Caltech101-20, \item VGGFace2-50, YTF. \end{itemize}  \\
\hline
\cite{Yuan2022} & 2022 &\begin{itemize}[noitemsep,topsep=0pt, left=0.1cm] \item[\checkmark] Graph-based \item[\checkmark] Kernel-based  \end{itemize}& \begin{itemize}[noitemsep,topsep=0pt, left=0.1cm] \item A robust self-tuning multi-view clustering method.  \item It solves the problems of existing multi-view clustering methods, such as initialization sensitivity, fixing the number of clusters, and reducing the influence of outliers. \end{itemize}& \begin{itemize}[noitemsep,topsep=0pt, left=0cm] \item Advertisement, 3source, Flowers \item Caltech101, ImageS, Cornell, \item Texas, Washingdon,  Wisconsion. \end{itemize} \\
\hline
\cite{Wang-Siwei2022} & 2022 &\begin{itemize}[noitemsep,topsep=0pt, left=0.1cm] \item[\checkmark] Graph-based \item[\checkmark] Subspace-based\item[\checkmark] Anchor-based  \end{itemize}& \begin{itemize}[noitemsep,topsep=0pt, left=0.1cm] \item Fast parameter-free multiview subspace clustering with consensus anchor guidance.  \item A subspace clustering method with linear time complexity, joint anchor selection and graph construction, and parameter-free characteristics for large-scale applications.\end{itemize} & \begin{itemize}[noitemsep,topsep=0pt, left=0cm] \item Caltech101-20, Caltech101-all, \item CCV, SUN-RGBD, NUS-WIDE, \item AWA, MNIST, YoutubeFace. \end{itemize}   \\
\hline
\cite{Kang2022}\footnote{See Section 6.5 for the formal definition of the method.} & 2021 &\begin{itemize}[noitemsep,topsep=0pt, left=0.1cm] \item[\checkmark] Graph-based \item[\checkmark] Anchor-based\item[\checkmark] Bipartite-graph  \end{itemize}& \begin{itemize}[noitemsep,topsep=0pt, left=0.1cm] \item A scalable graph learning framework that incorporates anchor points and the concept of a bipartite graph.  \item In contrast to conventional, a bipartite graph is constructed to illustrate the relationship between samples and anchor points.  \item A connectivity constraint is used to represent clusters directly by connected components. \end{itemize}& \begin{itemize}[noitemsep,topsep=0pt, left=0cm] \item Caltech101-7, \item Citeseer, \item NUS. \end{itemize}  \\
\hline
\cite{Hajjar2021}\footnote{See Section 6.3 for the formal definition of the method.} & 2021 &\begin{itemize}[noitemsep,topsep=0pt, left=0.1cm] \item[\checkmark] Graph-based \end{itemize}& \begin{itemize}[noitemsep,topsep=0pt, left=0.1cm] \item Multi-view spectral clustering via constrained nonnegative embedding, overcoming limitations of traditional spectral clustering by integrating constraints for smoother nonnegative embedding and orthogonal columns. \end{itemize}& \begin{itemize}[noitemsep,topsep=0pt, left=0cm] \item COIL201, ORL, Out-Scene,  \item BBCSport, Caltech101, MSRCv1, \item Extended-Yale, MNIST-10000. \end{itemize}   \\
\hline
\cite{Yu2021} & 2021 &\begin{itemize}[noitemsep,topsep=0pt, left=0.1cm] \item[\checkmark] Graph-based \item[\checkmark] Subspace-based  \end{itemize}& \begin{itemize}[noitemsep,topsep=0pt, left=0.1cm] \item An approach to similarity merging in multi-view spectral clustering that addresses the challenge of assigning uniform weights to all samples within a view.  \item It deals with biased or missing elements in incomplete views, and using sparse subspace clustering to form initial similarity matrices.\end{itemize}& \begin{itemize}[noitemsep,topsep=0pt, left=0cm] \item BBCSport, ORL, \item Still DB, MSRC, \item UCI, NUS, 3-Sources. \end{itemize}  \\
\hline
\cite{Horie2021} & 2021 &\begin{itemize}[noitemsep,topsep=0pt, left=0.1cm] \item[\checkmark] Graph-based \item[\checkmark] Subspace-based  \end{itemize}& \begin{itemize}[noitemsep,topsep=0pt, left=0.1cm] \item An approach to merge similarities in multi-view spectral clustering.  \item Addresses the challenge of assigning uniform weights to all samples within a view, manages missing elements in incomplete views, \item Uses sparse subspace clustering to form initial similarity matrices.  \end{itemize}& \begin{itemize}[noitemsep,topsep=0pt, left=0cm] \item BBC, \item NGs, \item WebKB, \item 100leaves. \end{itemize}  \\
\hline
\cite{Yin2021} & 2021 &\begin{itemize}[noitemsep,topsep=0pt, left=0.1cm] \item[\checkmark] Graph-based \item[\checkmark] Subspace-based\item[\checkmark] Kernel-based  \end{itemize}& \begin{itemize}[noitemsep,topsep=0pt, left=0.1cm] \item A one-step multi-view spectral clustering method that addresses the problem of inconsistency.  \item It splits the non-negative embedding matrix into two matrices: the joint non-negative embedding matrix, which represents the joint cluster structure, and the specific non-negative embedding matrix, which represents the specific cluster structure for each view.\end{itemize} & \begin{itemize}[noitemsep,topsep=0pt, left=0cm] \item COIL20, MSRC-v1, \item Caltech101-7, \item Caltech101-20, \item ORL, 3-Sources. \end{itemize} \\
\hline
\cite{Chen-Peng2021} & 2021 &\begin{itemize}[noitemsep,topsep=0pt, left=0.1cm] \item[\checkmark] Graph-based \item[\checkmark] Subspace-based\item[\checkmark] Graph filtering  \end{itemize}& \begin{itemize}[noitemsep,topsep=0pt, left=0.1cm] \item A smoothed multiview subspace clustering that preserves the geometric features of the graph through graph filtering, which simplifies the subsequent clustering process.\end{itemize} & \begin{itemize}[noitemsep,topsep=0pt, left=0cm] \item Handwritten, Citeseer \item Caltech101-7. \end{itemize}  \\
\hline
\cite{Niu2021} & 2021 &\begin{itemize}[noitemsep,topsep=0pt, left=0.1cm] \item[\checkmark] Graph-based \item[\checkmark] Subspace-based  \end{itemize}& \begin{itemize}[noitemsep,topsep=0pt, left=0.1cm] \item A one-step multiview subspace clustering method with incomplete views.  \item It uses low-rank matrix factorization to learn a consensus representation matrix, and then combines it with the objective function of non-negative embedding and spectral embedding subspace clustering.  \end{itemize} & \begin{itemize}[noitemsep,topsep=0pt, left=0cm] \item BBCSport, NGs, WebKB, \item BUAA, Orl, Yale, NUS-WIDE, \item Caltech101, CCV. \end{itemize}\\
\hline
\cite{Xie2020} & 2020 &\begin{itemize}[noitemsep,topsep=0pt, left=0.1cm] \item[\checkmark] Graph-based \item[\checkmark] Subspace-based  \end{itemize}& \begin{itemize}[noitemsep,topsep=0pt, left=0.1cm] \item A subspace learning based multiview clustering method.  \item It derives a joint latent representation from the latent subspace rather than from the original data space by linear transformation. \item  The latent representation has a low-rank structure, which reduces computational complexity. The similarity matrix is then dynamically learned from this latent representation using manifold learning.\end{itemize}& \begin{itemize}[noitemsep,topsep=0pt, left=0cm] \item MSRC-v1, \item UCI Digits, \item NUS-WIDE, \item Scene15. \end{itemize} \\
\hline
\cite{Kang2020} & 2020 &\begin{itemize}[noitemsep,topsep=0pt, left=0.1cm] \item[\checkmark] Graph-based  \end{itemize}& \begin{itemize}[noitemsep,topsep=0pt, left=0.1cm] \item An innovative multiview spectral clustering model capable of performing graph fusion and spectral clustering simultaneously.  \item The fusion graph approximates the original graph from each individual view while maintaining a clear and distinct cluster structure. \end{itemize}& \begin{itemize}[noitemsep,topsep=0pt, left=0cm] \item BBC, Reuters, \item Digits, Caltech101-20. \end{itemize}  \\
\hline
\cite{Wang-Hao2020} & 2020 &\begin{itemize}[noitemsep,topsep=0pt, left=0.1cm] \item[\checkmark] Graph-based   \end{itemize}& \begin{itemize}[noitemsep,topsep=0pt, left=0.1cm] \item The approach can jointly estimate the similarity graph matrix, the unified graph matrix, and the final cluster assignment by an innovative multi-view fusion technique.  \item This method imposes a rank constraint on the Laplacian matrix, ensuring the accurate derivation of clusters. \end{itemize}& Two toy data sets \begin{itemize}[noitemsep,topsep=0pt, left=0cm] \item Two-Moon, \item Three-Ring. \end{itemize}  \\

\hline
\cite{Hu2020} & 2020 & \begin{itemize}[noitemsep,topsep=0pt, left=0.1cm] \item[\checkmark] Graph-based  \end{itemize}&\begin{itemize}[noitemsep,topsep=0pt, left=0.1cm] \item A simple and efficient approach to multiview spectral clustering that aims to learn a sparse similarity matrix that is consistent across all views.  \item The advantage of the model lies in its ability to directly obtain a closed-form solution without the need for iteration.  \item The method introduces only one additional parameter, which depends on the way the similarity matrix is constructed and can be practically set to a small value. \end{itemize}& \begin{itemize}[noitemsep,topsep=0pt, left=0cm] \item Caltech101, \item MSRCv1, \item NUS-WIDE, \item ORL, \item 3-Sources. \end{itemize}   \\
\hline
\cite{Zhou2020} & 2020 &\begin{itemize}[noitemsep,topsep=0pt, left=0.1cm] \item[\checkmark] Subspace-based  \end{itemize}& \begin{itemize}[noitemsep,topsep=0pt, left=0.1cm] \item An approach that learns the joint information to leverage the underlying correlations across multiple views while capturing view-specific details to represent specific features for each independent view.  \item This is achieved without being affected by redundancy or the high dimensionality of the data.\end{itemize}& \begin{itemize}[noitemsep,topsep=0pt, left=0cm] \item Yale, MSRCV1, \item Caltech101-7, \item BBCSport, CMU-PIE. \end{itemize}  \\
\hline
\cite{Huang2019}\footnote{See Section 6.1 for the formal definition of the MVCSK method.} & 2019 &\begin{itemize}[noitemsep,topsep=0pt, left=0.1cm] \item[\checkmark] Graph-based \item[\checkmark] Kernel-based  \end{itemize}& \begin{itemize}[noitemsep,topsep=0pt, left=0.1cm] \item An innovative multiview learning model is presented that is capable of simultaneously performing a multiview clustering task and capturing similarity relationships in kernel spaces.  \item The model autonomously assigns optimal weights to each view without the need for additional parameters.\end{itemize}& \begin{itemize}[noitemsep,topsep=0pt, left=0cm] \item Texas, Cornell, \item Washington	\item Winconsin, \item BBC, BBCSport,  NUS-WIDE.\end{itemize} \\
\hline
\cite{Chao2019} & 2019 &\begin{itemize}[noitemsep,topsep=0pt, left=0.1cm] \item[\checkmark] Co-clustering  \end{itemize}& \begin{itemize}[noitemsep,topsep=0pt, left=0.1cm] \item The method addresses the problem of missing data in multiview clustering. In contrast to existing multiview co-clustering approaches that struggle with incomplete data, especially when different patterns of missing data are present. \item This method uses an indicator matrix. The indicator matrix highlights which data elements are present and clustering performance is measured solely on the observed values.\end{itemize} & \begin{itemize}[noitemsep,topsep=0pt, left=0cm] \item Clinical data from the heroin treatment study.\end{itemize} \\
\hline
\cite{Wang-Hao2019} & 2019 &\begin{itemize}[noitemsep,topsep=0pt, left=0.1cm] \item[\checkmark] Graph-based \item[\checkmark] Consensus learning\item[\checkmark] Perturbation risk \end{itemize}& \begin{itemize}[noitemsep,topsep=0pt, left=0.1cm] \item Addresses clustering of multi-view data with missing instances using spectral perturbation theory to construct similarity matrices and learn consensus Laplacian matrix. \end{itemize}& \begin{itemize}[noitemsep,topsep=0pt, left=0cm] \item 100Leaves, Flowers17, Mfeat, \item ORL, 3Sources, BBCSport.\end{itemize} \\
\hline
\cite{Peng2019} & 2019 &\begin{itemize}[noitemsep,topsep=0pt, left=0.1cm] \item[\checkmark] Graph-based   \end{itemize}& \begin{itemize}[noitemsep,topsep=0pt, left=0.1cm] \item A method that estimates the number of clusters by providing cross-view consensus on view-specific similarity graphs instead of relying on view-specific data representations.  \item  this end, a novel objective function is used to project the raw data into a space where the projection accounts for geometric consistency and cluster assignment consistency. \end{itemize}&\begin{itemize}[noitemsep,topsep=0pt, left=0cm] \item Caltech101, \item LandUse-21, \item Scene-15, \item Still-DB.\end{itemize}   \\
\hline
\cite{Xing2019} & 2019 &\begin{itemize}[noitemsep,topsep=0pt, left=0.1cm] \item[\checkmark] Subspace-based \end{itemize}& \begin{itemize}[noitemsep,topsep=0pt, left=0.1cm] \item A correntropy-based method for dealing with noise in multiview clustering using a view-specific embedding from an information-theoretic perspective.  \item The objective function uses the Frobenius norm to efficiently estimate the dense connections between points that lie in the same subspace.\end{itemize}& \begin{itemize}[noitemsep,topsep=0pt, left=0cm] \item UCI Digits, 3-Sources, \item Movies617, BBC, \item Cora, Washington.\end{itemize}    \\
\hline
\cite{Zhan2018} & 2018 &\begin{itemize}[noitemsep,topsep=0pt, left=0.1cm] \item[\checkmark] Graph-based \end{itemize}& \begin{itemize}[noitemsep,topsep=0pt, left=0.1cm] \item A graph learning framework, optimizing initial graphs from different views using low-rank constraint Laplacian matrix.  \item It leads to a global unified graph estimation. \end{itemize}& \begin{itemize}[noitemsep,topsep=0pt, left=0cm] \item UCI Digits, \item Caltech101, \item Notting-Hill, COIL-20.\end{itemize}  \\
\hline
\end{longtable}

\end{landscape}

\section{Classical Methods for Multi-view Clustering}
\label{Classical-methods}

\begin{figure}[h!]
\begin{center}
	\includegraphics[width=.8\textwidth]{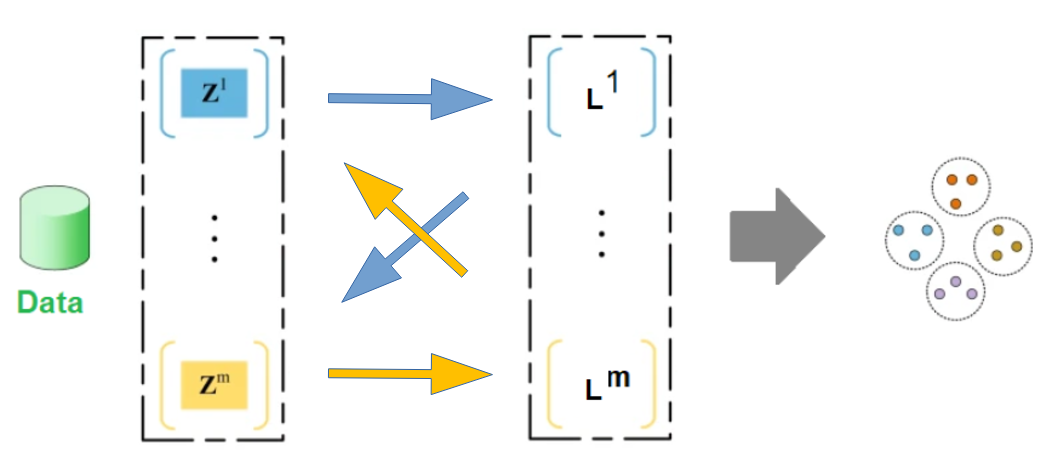}
	\caption{A graphic illustration of the co-training scheme with several views. The matrices $\{\Zvect^1,...,\Zvect^m\}$ refer to the different data matrices corresponding to each view and $\{\Lvect^1,...,\Lvect^m\}$ are the co-trained models. In this scheme, the information obtained from the individual views is systematically refined and iteratively exchanged, which promotes joint learning from the different views.} \label{fig:co-traingmvc}
\end{center}
\end{figure}

Classical machine learning approaches for multi-view learning encompass diverse strategies. The co-training approach, as proposed by Blum et al. \cite{Blum1998}, uses multiple data views to improve model performance. Each view is trained independently and the information between the views is iteratively exchanged and refined. Figure \ref{fig:co-traingmvc} shows a graphical illustration of this idea.

Co-training is particularly beneficial in scenarios where there is little labeled data and allows the effective use of unlabeled data. However, practical considerations such as selecting features, addressing class imbalances and defining a robust matching criterion for the selection of instances to be labeled are essential. Co-training has applications in various fields, including natural language processing, computer vision and bioinformatics.

The co-regularized approach for semi-supervised learning from multiple views presented by Kumar et al (\cite{Kumar2011}) provides a powerful technique. This method combines the principles of co-regularization and semi-supervised learning to jointly learn representations for each view and make predictions in an unsupervised manner. Co-regularization enforces a constraint that ensures that the representations of the same instance from different views are close to each other, which promotes robust and discriminative representations.

Multi-kernel spectral clustering (MKSC) integrates kernel features from different views into a unified kernel matrix and uses spectral clustering for effective multi-view clustering \cite{Xia2014}. Canonical correlation analysis (CCA) is another approach that identifies correlated subspaces between views, as described in \cite{Chaudhuri2009,Feng2022}.

Alternative strategies include the Joint Non-Negative Matrix Factorization approach, where data matrices from different views are factorized into non-negative matrices to capture common and view-specific information. Li et al. \cite{Li2018} demonstrate spectral clustering on the learned representations for multi-view clustering.

\begin{figure}[h!]
	\includegraphics[width=1\textwidth]{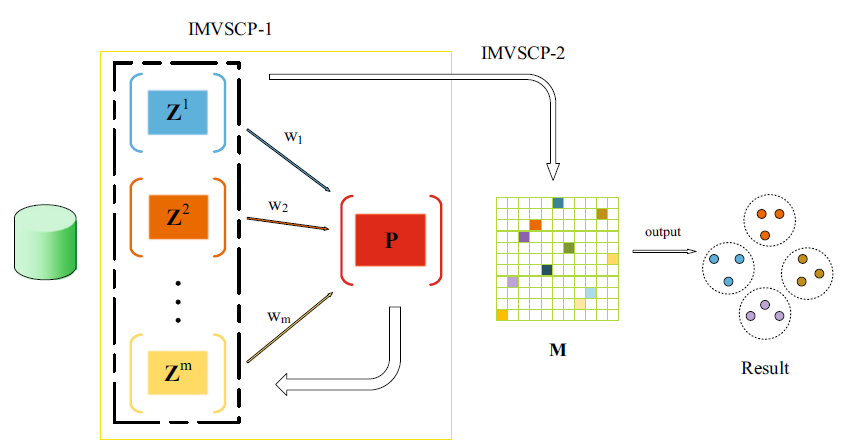}
	\caption{The flow chart of the Multi-view Spectral Clustering  \cite{Chen2022}. First, similarity matrices of the individual views ($\Zvect$) are created using the K-nearest neighbor algorithm. Then, all views are combined into a unified matrix $\Pvect$ using a weighted fusion operation. This unified matrix $\Pvect$ is used to update the similarity matrix for each view. The algorithm iteratively refines the similarity matrices obtained in the previous step. It integrates the spectral clustering algorithm and the symmetric non-negative matrix factorization algorithm to generate the non-negative embedding matrix $\Mvect$, which leads directly to the clustering results.}\label{fig:mvspc}
\end{figure}

Multi-View Spectral Embedding (MVSE) constructs individual similarity graphs for each view and merges them into a unified graph, which is subjected to spectral embedding for low-dimensional representations and subsequent clustering \cite{Xu2019}. Chen et al. \cite{Chen2022} propose an improved spectral multiview clustering method that incorporates tissue-like P-systems and optimizes the similarity matrices through a weighted iterative process. The approach combines the K-nearest neighbor algorithm, a weighted fusion operation, and an iterative update to obtain non-negative embedding matrices and achieve clustering results.  Figure \ref{fig:mvspc} shows the flow of chart of this approach. The method first determines the similarity matrices of each view ($\Zvect$) by the K-nearest neighbor algorithm and then fuses all views into a unified matrix $\Pvect$ by a weighted fusion operation. The unified matrix $\Pvect$ in turn updates the similarity matrix for each view. Through an iterative updating process, the algorithm then takes the updated similarity matrices obtained in the previous step as input and combines the spectral clustering algorithm and the symmetric non-negative matrix factorization algorithm to obtain the non-negative embedding matrix $\Mvect$ to directly output the clustering results.

Recent advances include joint frameworks that incorporate the orthonormality constraint and co-regularization \cite{Cai2020}. This improves class differences and feature scale consistency, which contributes to improved multiview clustering performance. In addition, Hajjar et al. \cite{Hajjar2021} constraints to maintain consistent smoothing and enforce orthogonality, which improves the robustness and stability of clustering results. This method overcomes the limitations of conventional spectral clustering and provides more reliable and accurate results.

\subsection{Kernel-Based Approaches}
\label{kernel-based-methods}
Multi-view spectral clustering algorithms are designed to effectively handle data with multiple views, exploiting the complementary information present in each view to improve the quality of clustering results. In their pursuit of optimal performance, many {\bf multikernel-based methods} have also been developed with the main goal of obtaining a unified kernel by specifying a predefined kernel for each different view and then combining all kernels linearly or nonlinearly.

\begin{figure}[h!]
	\includegraphics[width=1\textwidth]{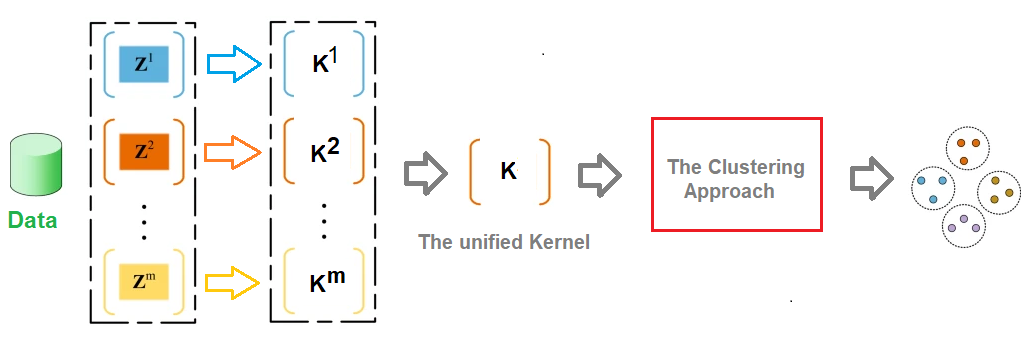}
	\caption{The flowchart of kernel-based multi-view spectral clustering. For multi-view data, we denote ($\Zvect^1,\Zvect^2,...,\Zvect^m$) as the data matrix with $m$ views. Then an individual kernel is constructed for each view so that we have ($\Kvect^1, \Kvect^2,...,\Kvect^m$). A specific clustering method is then selected to cluster the data based on the unified kernel $\Kvect$. }\label{fig:mvspc-kernel}
\end{figure}

Kernel functions are an essential part of these methods. They define a measure of similarity or affinity between data points in the transformed feature space. Commonly used kernel functions include the linear kernel, which computes the dot product between data points in the original feature space and thus measures their linear similarity, and the radial basis function (RBF) kernel, which measures similarity based on the Euclidean distance between data points in the transformed space and thus captures nonlinear relationships. Figure \ref{fig:mvspc-kernel} gives a graphical illustration of the flow chart of Kernel-based approaches for multi view clustering.

For example, in \cite{Huang2019}, the authors discuss two kernel-based graph learning multi-view clustering methods with automatic weights. These two methods are used to map the data into a space where they are linearly separable. The first method uses a single kernel per view, while the second method uses a combination of multiple kernel matrices to improve the utilization of the input kernel matrix. Moreover, these methods simultaneously estimate the unified similarity matrix, the consistent spectral projection matrix, and the weight of each view without additional parameters.

Many of these approaches follow a two-step process to achieve clustering results. The initial step involves learning the joint affinity matrix, while the second step utilizes a hard clustering method like k-means clustering to obtain the final result. To address the inconsistency issue arising from the fact that the primary objective of the first step is not optimal clustering performance, a novel method is introduced.

In \cite{Zhu2018}, the authors present a method called One-step Multi-view Spectral Clustering (OMSC), which integrates the steps of learning the affinity matrix of each view and the joint affinity matrix learned from the low-dimensional space of the data, as well as the step of k-means clustering in one framework. The joint affinity matrix is considered as the final clustering assignment. Moreover, the weighting of each view is automatically learned to reduce the impact of noisy views.

In \cite{Ren2020}, the authors jointly estimate an optimal graph and an appropriate consensus kernel for clustering by forcing the global kernel matrix to be a convex combination of a set of basis kernels. Their proposed model enforces a regularization of the unified graph and the final kernel matrix.

However, the performance of multikernel-based methods strongly depends on the predefined kernel, including the kernel type (e.g. Gaussian kernel, linear kernel and polynomial kernel) and the corresponding parameters. In addition, most of these methods do not exploit the label space of the data (i.e. the soft mappings of the cluster) and only extract information from the data space. New approaches have been developed to utilise the information in the label space. For instance, in \cite{Hajjar2022}, the authors develop a new method that incorporates the non-negative embedding matrix, which can be used as a cluster indicator matrix to perform the final cluster assignment without further post-processing steps such as k-means or spectral rotation. Unlike other multi-view clustering methods, this method can create a new graph based on the soft cluster labels. Moreover, this method smoothes the cluster-label indices over the data graphs and the label graph, which improves the performance.

\subsection{Subspace-Based Approaches}
\label{subspace-based-methods}

Subspace clustering has been extensively researched over the last ten years due to its promising performance. Basically, it assumes that the data points come from multiple low-dimensional subspaces, with each cluster fitting into one of these low-dimensional subspaces. Considerable progress has been made in uncovering these underlying subspaces. In this context, several innovative methods have been proposed to overcome the challenges associated with multi-view data. In this section, four notable works are presented, each of which presents a different approach for clustering subspaces from multiple perspectives.

The work of Zhang et al. \cite{Zhang-Changqing2017} provides a method called Latent Multi-view Subspace Clustering (LMSC). LMSC introduces a unique perspective by clustering data points with latent representations while exploring complementary information from multiple views. Unlike traditional single-view subspace clustering, LMSC searches for an underlying latent representation, resulting in more accurate and robust subspace representations. The method is intuitive and is efficiently optimized by the Augmented Lagrangian Multiplier with Alternating Direction Minimization (ALM-ADM) algorithm, which has been confirmed by extensive experiments on benchmark datasets.

Khan et al. \cite{Khan2023}, however, addresses a limitation in existing multiview subspace clustering approaches by considering the structure of self-representation through consistent and specific representations. Based on low-rank sparse representations, the method uncovers global common representation structures among views and preserves geometric structures based on consistent and specific representations. 

Other \cite{Khan2021} focuses on integrative clustering for high-dimensional data with multiple views based on manifold optimization. To identify consensus clusters, the algorithm constructs a joint graph Laplacian and optimizes a joint clustering target while minimizing the disagreements between individual and joint views. This optimization is alternatively performed over k-means and Stiefel manifolds \cite{james1976topology}, modeling nonlinearities and differential clusters within individual views. The convergence of the algorithm is demonstrated over the manifold, and experimental results on benchmark and multi-omics cancer datasets show its superiority over existing multi-view clustering approaches.

The work of \cite{Wang-Siwei2022} addresses the efficiency challenges in multi-view subspace clustering by proposing a Fast Parameter-free Multi-view Subspace Clustering with Consensus Anchor Guidance (FPMVS-CAG). This method performs anchor selection and subspace graph construction in a unified optimization formulation, thus promoting clustering quality. FPMVS-CAG has linear time complexity with respect to the number of samples and automatically learns an optimal anchor subspace graph without additional hyperparameters.


\subsection{Integration of Deep Learning with Multi-view Clustering}
\label{Deep-methods}

\modified{Integrating deep learning with multi-view clustering is a modern approach that uses neural networks to create joint representations for different views of data. This strategy improves clustering performance by learning feature representations that capture both intra-view and inter-view relationships. However, despite the growing interest and proliferation of algorithms based on various theories, many existing models are superficial and map multi-view data directly to low-dimensional spaces. }

\modified{ To address the limitation of shallow models, deep multi-view clustering algorithms have emerged as a promising solution. Du et al. \cite{Du2021deep} propose a methodology utilizing multiple autoencoders with a layer-wise approach to capture nonlinear structural information within each view. This model incorporates local invariance within a view and integrates consistent, complementary information between views. The algorithm surpasses shallow models by unifying representation learning and clustering in a cohesive framework, enabling joint optimization of both tasks. }

\modified{ In a similar context, \cite{Wang2024} combines the flexibility of deep learning with the statistical benefits of data-driven and knowledge-driven feature selection, providing interpretable results. It learns nonlinear relationships in multi-view data by using deep neural networks to create low-dimensional, view-independent embeddings, while imposing a regularization penalty on the reconstructed data. The method uses the normalized Laplacian of a graph to model bilateral relationships between variables within each view, promoting the selection of related variables. }

\added{ Other approaches \cite{Wang-Jiao2024} attempt to overcome the limitations of traditional deep multi-view subspace clustering approaches by using a decomposed optimization strategy that involves three stages: Pre-training with auto-encoders to extract multiscale features, fine-tuning to learn consensus self-expressions and generate high-quality pseudo-labels, and re-training with self-label supervision for robust clustering. This approach improves clustering performance by addressing issues such as skew in pipelined methods and complex parameter optimization in end-to-end methods.}

\modified{
\begin{itemize}
    \item \textbf{Deep subspace approaches:}  Multiview subspace clustering (MVSC) uses complementary information from multiple views to achieve better clustering compared to single-view methods. However, it often struggles with high-dimensional and noisy raw data. To overcome this, the self-guided deep multiview subspace clustering (SDMSC) model \cite{Li-Kai2022} combines deep feature embedding with subspace analysis to uncover a reliable consensus data affinity relationship across views and embedding spaces. By using raw feature affinities as monitoring signals, SDMSC controls the embedding process itself, reduces the risk of poor local minima, and improves clustering performance. Similarly, Zhu et al. \cite{Zhu2024} address the major limitations of traditional MVSC methods, such as insufficient multiview integration and lack of end-to-end learning, with the Multiview Deep Subspace Clustering Network (MvDSCN). This approach uses two sub-networks: Dnet for view-specific representations and Unet for joint representations. Deep convolutional autoencoders are used to construct a multiview self-representation matrix in an end-to-end method. 
    \item \textbf{Advances in graph neural networks:} Recent advancements in graph neural networks (GNNs) have provided new methods for multi-view learning. GNNs have garnered significant attention in recent years \cite{Wu2021,jiang2023transfer,rebafka2024model}, with their primary idea being the embedding of node representations by capturing and aggregating information from local neighborhoods. This technique has shown great promise, especially in tasks such as social network analysis, recommendation systems, anomaly detection, and the analysis of physiological data \cite{liu2018heterogeneous,wu2019session,pucci2006investigation,GRANA2023126901}. GNNs integrate information from multiple views, either through graph structures or self-supervised learning, leading to improved predictions at both the node and graph levels, and efficiently utilizing unlabeled data.
    \item \textbf{Generative models for multi-view learning:} In parallel, generative models such as Variational Autoencoders (VAEs) and Generative Adversarial Networks (GANs) have been adapted to handle multi-view data \cite{zhang2023multi,goodfellow2014generative,Wang-Yang2018}. These models enable the generation of data samples from multiple views, offering advantages such as data augmentation and enhanced model robustness. Adversarial multi-view learning techniques, in particular, align representations across different views while preserving view-specific characteristics, thus addressing the challenges posed by view heterogeneity. For instance, Tang et al. \cite{Tang2023} propose Consistent and Diverse Deep Latent Representations (CDDLR), which integrates K-means with spectral clustering to enforce consistency and diversity in the latent representations. This approach employs Laplacian-regularized deep neural networks to maintain diversity and consistency, providing a robust solution for multi-view subspace clustering.
    \item \textbf{Solutions for deep multi-view subspace clustering:} To address challenges in deep multi-view subspace clustering, Shi and Zhao \cite{Shi2023} introduce a novel solution known as Deep Multi-View Clustering based on Reconstructed Self-Expressive Matrix (DCRSM). This approach utilizes a reconstruction module to approximate self-expressive coefficients and resolves scalability issues by employing a minimal number of training samples. It effectively combines common and specific layers, facilitating the fusion of consistent and view-specific information, leading to improved clustering outcomes. Additionally, Wang et al. \cite{Wang-Jiao2024} propose Decomposed Deep Multi-View Subspace Clustering with Self-Labeling Supervision (D2MVSC), which addresses traditional challenges in deep multi-view subspace clustering through a decomposed optimization strategy. This method includes a three-stage training process, adaptive fusion, and structure supervision, improving clustering accuracy and robustness.
    \item \textbf{Further contributions to multi-view learning:} Wang et al. \cite{Wang20234371} contribute to subspace-based multi-view learning by proposing Multi-view Orthonormalized Partial Least Squares (MvOPLSs). This framework incorporates regularization techniques for model parameters, decision values, and latent projected points. By extending the framework with nonlinear transformations using deep networks, MvOPLSs achieves superior performance across diverse multi-view datasets compared to existing methods. These innovations mark significant progress in deep multi-view subspace clustering and contribute to the ongoing evolution of multi-view learning techniques.
\end{itemize}
Recent advancements in multi-view learning underscore the growing potential of multi-view data to enhance model performance, robustness, and generalization. The choice of algorithm often depends on the specific characteristics of the data and the nature of the problem. Approaches such as deep learning and generative techniques provide comprehensive solutions, enabling the effective handling of complex multi-view data and improving clustering outcomes across a variety of applications.
}

\section{Exploring Graph-Based Multi-View Clustering}
\label{summary-table}

Graph-based multi-view clustering, an essential aspect of data analysis, has made significant progress with the introduction of various methods. This section provides an overview of the latest approaches developed over the last five years, summarizing key contributions, including novel algorithms and methods that address the main challenges. These contributions provide valuable insights into the evolving landscape of multi-view clustering, offering significant contributions to the broader field of machine learning and data analytics. Researchers working on multi-view clustering will find this compilation valuable for understanding and navigating these developments.
\begin{figure}[h!]
	\includegraphics[width=1\textwidth]{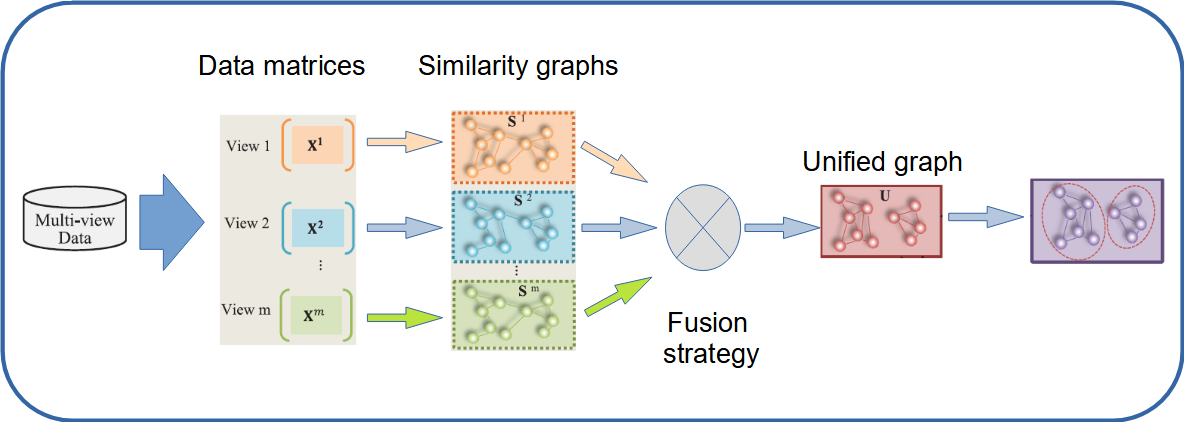}
	\caption{The flowchart of graph-based multi-view clustering, adapted from \cite{Wang-Hao2020}. In this scheme, the data matrix of each view is converted into a graph matrix, followed by applying a fusion method across all views to create a unified graph.}\label{fig:graphmvc}
\end{figure}
\modified{
\begin{itemize}
    \item \textbf{Multi-view graph clustering:} The goal is to find a fusion graph across all views and apply graph-cut algorithms (e.g., spectral clustering) on the fusion graph to produce the final clustering result. We focus on graph-based clustering, where each view’s data points are represented by a graph. These graphs are combined to facilitate clustering of the entire dataset, utilizing complementary information from different views to improve cluster mapping and provide a more comprehensive understanding of complex data structures.   
    \item \textbf{Schematic representation of graph-based methods:} Figure \ref{fig:graphmvc} illustrates the process. Here, the data matrix from each view is converted into a graph matrix, followed by a fusion method applied to create a unified graph. The fusion process derives the weights for each view automatically using a novel multi-view fusion technique. The unified graph similarity matrix is learned jointly with the individual graph similarity matrices, representing pairwise similarities between data samples.    
 \item \textbf{Limitations and Challenges in Graph-based Multi-view Clustering:} Traditional graph-based methods face several challenges that impact their performance. Issues such as sample selection bias, additional clustering steps, and variations in similarity metrics complicate the clustering process \cite{Wang-Hao2020}. Furthermore, specific challenges in graph-based methods include:
    \begin{itemize}
        \item \textbf{Post-processing necessity:} Final clustering results often require methods like K-Means or spectral rotation to achieve consistent spectral embeddings, introducing uncertainties due to initialization.
        \item \textbf{Parameter selection:} Many graph-based models introduce additional parameters, complicating the task, which is inherently unsupervised.
        \item \textbf{High computational costs:} Spectral multi-view clustering methods require eigenvalue decomposition with a computational complexity of $O(n^3)$, and multi-view subspace clustering requires matrix inversion with similar computational complexity, making these methods particularly challenging for large-scale data \cite{HU2020251}.
    \end{itemize}
    Approaches like ASMV (Adaptive Similarity Metric Fusion for Multi-view Clustering) \cite{Zhang2017} and GMC (Graph-based Multi-view Clustering) \cite{Wang-Hao2020} have been proposed to mitigate these limitations and improve the effectiveness of graph-based multi-view clustering.    
    \item \textbf{A novel solution - MCGLSR:} In response to these challenges, a new method called \emph{Single Phase Multi-view Clustering using Unified Graph Learning and Spectral Representation (MCGLSR)} \cite{Dornaika2023} is proposed. Unlike conventional methods that directly integrate similarity matrices (which may introduce noise), MCGLSR generates similarity graphs and a joint similarity matrix using a unified global objective function. This ensures that similarity matrices from different views align effectively, mitigating noise and promoting a more coherent data structure. The method outputs individual graphs, a joint graph, a joint spectral representation, and cluster mappings, eliminating the need for an external clustering algorithm.  
    \item \textbf{Incomplete data handling:} In the next section, we address the challenges of multi-view clustering with incomplete data. \cite{Li2024} presents a breakthrough solution to the limitations of subspace clustering in the context of incomplete data. This approach uses a multiple kernel completion scheme to detect intact kernels and ensures the learning of complete, low-redundancy representations. By integrating multi-view subspaces with a weighted tensor low-rank constraint, the method explores higher-order relationships between views and assigns appropriate weights to each view. The result is a unified model that learns low-redundancy representations, view-specific subspaces, and their low-rank tensor structure, significantly improving subspace clustering in incomplete data scenarios.
\end{itemize}
}

\section{Multi-View Clustering With Missing or Incomplete Data}
\label{mvc-missing-data}

The rapid development of technology has led to a massive growth in data, with multi-view data offering complementary information from different sources. This makes multi-view data more informative than single-view data. However, in real-world applications, multi-view data is often incomplete. For example, a video without sound or data points with missing features represent incomplete views. Traditional multi-view clustering techniques cannot handle such incomplete data \cite{Wen2022SurveyIM}.

Dealing with incomplete multiview data has become an important research topic that has attracted much attention in recent years \cite{wen2020adaptive, xia2022tensor, diallo2023auto}. Several methods have been proposed to solve this problem:
\modified{
\begin{itemize}
\item \textbf{Filling missing views:} A common approach is to fill missing views with zeros or the average values of the available instances for that view \cite{shao2016online, shao2015multiple}. In addition, weighted non-negative matrix factorization with $\ell_{2,1}$ regularization was used to improve clustering performance on incomplete views. However, this method often results in clustered data that lacks meaningful information, as filled values may cluster unrelated samples together.
\item \textbf{Consensus-based anchor guidance:} A more reasonable approach proposed in \cite{Wang-Siwei2022} emphasizes the use of available information in incomplete multiview data. This method uses consensus-based anchor guidance to achieve faster and more robust clustering, especially in image processing applications.
\item \textbf{Nonnegative matrix factorization (NMF):} In \cite{li2014partial}, NMF was used to construct a latent subspace from the available multiview data, ensuring correct alignment between the corresponding instances across the different views. In \cite{zhao2016incomplete}, graph embedding techniques were applied to capture global structure in heterogeneous data with missing values and improve clustering of multimodal visual data.
\item \added{\textbf{Hypergraph-based approach:} To address the challenges of processing incomplete multi-view data, the authors in \cite{Chen2025} propose a hypergraph-based approach. In contrast to traditional methods that focus on pairwise relations, their approach captures the higher order relations in the data using a hypergraph. This enables a more comprehensive exploration of both local and global data structures. Furthermore, the method combines non-negative matrix factorization with orthogonality constraints and K-means clustering, eliminating the need for post-processing.}
\end{itemize}
}
Apart from these approaches, additional techniques such as sparsity constraints, weighted learning, graph learning, local embedding structures, and global embedding structures have been integrated into Incomplete Multi-View Clustering (IMVC) models to further improve clustering performance \cite{zhou2023self, yuan2021adaptive, wang2019study}.

Despite the success of these techniques, there are still challenges to be addressed. Many models treat the contributions of all views equally, ignoring the varying discriminative information each view may provide. Additionally, IMVC methods are prone to misclassification, often resulting in fewer clusters than actually exist. Furthermore, there is a risk of imbalanced clustering, where some clusters are over-partitioned, while others are under-represented, even when the data is balanced.

The challenge of clustering multi-view data with missing information remains an active area of research. Future work should refine existing methods, explore new techniques, and address specific application areas to enhance the handling of incomplete multi-view data.

\section{Formal Review of Typical Approaches}
\label{ReviewTypicalApproach}
\secondround{Graph-based multi-view clustering (MVC) methods have made significant progress in recent years and have overcome important challenges in this area. The main ones are constructing reliable similarity graphs, mastering the computational complexity of large datasets, ensuring consistency between different views, and optimizing the integration of information from multiple views.\\
A critical problem with MVC is the construction of the similarity graph for each view, which often requires large amounts of memory and computational resources. Many methods have developed more efficient techniques for constructing graphs, either by integrating multiple views into a common framework or by using scalable algorithms that reduce computational costs without sacrificing accuracy.\\
Another major challenge is effectively assigning appropriate weights to views, which may have different importance in different datasets. Several methods have addressed this problem by proposing automatic or adaptive weight assignment mechanisms that eliminate the need for manual tuning and improve the robustness of the clustering process.\\
Consistency across multiple views is another key concern in MVC. Since different views may represent different perspectives or aspects of the same data, it is critical that the clustering results remain consistent across views. Newer approaches address this issue by explicitly modeling both consistent and inconsistent information, improving the overall accuracy and reliability of the clustering process.\\
Finally, the integration of various components involved in clustering, such as similarity matrices, spectral representations, and soft cluster assignments, has traditionally been handled sequentially. However, newer methods take a more integrated approach by optimizing these components together in a single framework, thus improving the efficiency and consistency of clustering results.\\
By overcoming these challenges, the methods described in the following sections contribute to scalable, robust and effective multi-view clustering and push the boundaries of what is possible with large, complex datasets.}

To begin, we introduce the primary notation used in this section, where matrices are shown in bold capital letters and vectors in bold lowercase letters. \\

Let $\Xvect^{(v)}$ be a data matrix $(\xvect_1^{(v)},\xvect_2^{(v)},...,\xvect_n^{(v)}) \in \mathbb{R}^{n \times d^{(v)}}$ where $n$ is the number of data instances, $d_v$ the number of features in view $v$ where $v= 1,...,V$.  Given a matrix $\Avect$, its trace is denoted by $\textit{Tr(\Avect)}$ and its transpose by $\Avect^T$. $A_{ij}$ is an element of the matrix $\Avect$.
 
\subsection { Auto-weighted Multi-view Clustering via Kernelized graph learning (MVCSK)}
\label{7-1}

\added{Graph-based approaches have been widely adopted for multi-view clustering due to their ability to reveal hidden structures within data. However, a major challenge in these methods is the construction of accurate similarity graphs, which can be influenced by several factors, including the scale of the data, neighborhood size, choice of similarity metric, and the presence of noise and outliers.\\
To overcome these limitations, \cite{Huang2019} propose a novel approach to multi-view clustering that simultaneously performs the clustering task and learns similarity relationships in kernel spaces. The key feature of this method is its ability to construct a similarity graph that can be directly partitioned into \( c \) connected components, where \( c \) is the number of clusters. This ensures that the clustering process is more accurate and aligned with the underlying data structure.\\
One of the most significant innovations of this approach is its automatic weight assignment for each view. Unlike traditional methods, which either require manual assignment of weights or introduce additional parameters, the model in \cite{Huang2019} learns the optimal weight for each view as part of the clustering process. This automatic weighting improves the robustness and adaptability of the clustering method, as it allows the model to adjust to the unique characteristics of the data without requiring prior knowledge of the view importance.\\
Additionally, the proposed model incorporates multiple kernel learning to address the sensitivity to the input kernel matrix. The inclusion of this extension enables the model to better handle nonlinear relationships in the data and ensures that it can capture complex patterns that would otherwise be missed by traditional graph-based clustering methods.}

\added{The model operates within a joint learning framework, which simultaneously solves three subtasks:
\begin{itemize}
    \item Constructing the most accurate similarity graph,
    \item Automatically assigning optimal weights to each view,
    \item Finding the cluster indicator matrix.
\end{itemize}
By solving these subtasks jointly, each task enhances the others, leading to a more robust and effective clustering outcome. Experimental results on benchmark datasets show that the method proposed by \cite{Huang2019} outperforms several state-of-the-art multi-view clustering algorithms, demonstrating its effectiveness in handling complex, incomplete, and noisy multi-view data.}

The objective function of MVCSK is:

\begin{equation}
\label{MVCSK}
\begin{split}
      &\min_{\Svect,  \; \Pvect}  \sum_{v=1}^V   \sqrt {Tr \, (\Kvect_v\,-2 \, \Kvect_v \, \Svect \,+\Svect^T \, \Kvect_v \, \Svect ) } + \mu \, ||\Svect||^{2}_2   
  +\lambda \, Tr \, (\Pvect^T \, \Lvect \, \Pvect) 
 \;\\
 &s.t.   \;   \Svect \ge 0,  \;  \Pvect^T  \Pvect = \Ivect,
\end{split}   
\end{equation}
where $\Svect$ is the unified similarity matrix, $\Kvect_v$ is the kernel matrix of each view, $\Pvect$ is the unified spectral projection matrix from which the final clustering is estimated using an extra step. 

\added{The objective function is designed to achieve two main goals: (1) to construct an accurate and smooth similarity matrix \(\Svect\) that reflects the true relationships between data points, and (2) to learn the optimal projection matrix \(\Pvect\) that enables the effective clustering of the data. The first term in the objective function minimizes the distance between the kernel matrices of each view and the unified similarity matrix, promoting consistency between the different views. The second term regularizes \(\Svect\) by enforcing its smoothness, while the third term regularizes the projection matrix \(\Pvect\) by encouraging it to preserve the underlying graph structure as encoded in the Laplacian matrix \(\Lvect\).\\
The constraints \( \Svect \ge 0 \) and \( \Pvect^T \Pvect = \Ivect \) ensure that the similarity matrix is non-negative and that the projection matrix is orthonormal, respectively. Together, these terms and constraints guide the model toward a robust and effective clustering solution.}

\subsection{Consistency-Aware and Inconsistency-Aware Graph-Based Multi-View Clustering (CI-GMVC)}
\label{7-2}

\modified{In \cite{Horie2021}, the authors propose a novel graph-based multi-view clustering method called CI-GMVC, which addresses a major limitation of existing approaches, including GMVC. Although conventional methods use a unified graph matrix for clustering multi-view data, they do not consider the inconsistent parts of the input graph matrices. This omission can lead to suboptimal clustering performance. The CI-GMVC method explicitly separates the consistent and inconsistent parts of the graph matrices and thus improves the robustness and accuracy of the clustering process when analyzing multi-view data.}

The proposed objective function aims to jointly estimate the consensus graph, the spectral representation of the data, and  the consistent graphs in each view. This is given by:

\begin{equation}
\begin{split}
     &\min_{\Avect_v=1,...,V, \; \Uvect, \;   \Fvect,\; \alpha}  \sum_{v=1}^V  \alpha_v \, ||\Uvect - \Avect_v||^2_2 \,+ \,2 \lambda \,{Tr \,\left( \Fvect^T \, \Lvect_U\,\Fvect \right)} \,
     + \sum_{v,w=1}^V  b_{vw} \, {Tr \,\left( (\Svect_v - \Avect_v) (\Svect_w - \Avect_w)^T)\right)} \\
     &\mbox{subject to} \; \; U_{ij} \ge  0, \quad \sum_j U_{ij} = 1, \quad \Fvect^{T} \Fvect = \Ivect,\quad  \Svect_{v} \ge   \Avect_v  \ge  0 
\end{split}
\end{equation}
where $\Svect_v$ is the graph matrix of the $v$-th view (known input), $\Uvect$ is the consensus graph, $\Fvect$ is the common spectral representation of the data, and $\Avect_v$ is the consistent graph matrix of view $v$.

\added{The objective function consists of several key terms, each with its own role in the clustering process:
\begin{itemize}
    \item \(\alpha_v ||\Uvect - \Avect_v||^2_2\): This term measures the difference between the consensus graph \(\Uvect\) and the consistent graph matrix \(\Avect_v\) for each view \(v\). The parameter \(\alpha_v\) controls the importance of this term for each view. Minimizing this term encourages the consensus graph to be close to the consistent graph for each individual view, ensuring that the consensus graph captures the consistent parts of the multi-view data. This helps to identify the parts of the data that are reliably shared across all views.    
    \item \(2 \lambda \, Tr(\Fvect^T \, \Lvect_U \, \Fvect)\): This term regularizes the common spectral representation \(\Fvect\) by enforcing smoothness on the graph defined by \(\Lvect_U\), the graph Laplacian associated with the consensus graph. The parameter \(\lambda\) controls the strength of the regularization, encouraging the spectral representation to preserve the structure of the consensus graph. This helps to capture the global structure of the data and ensures that the learned representation aligns with the consensus information across all views.    
    \item \(\sum_{v,w=1}^V b_{vw} \, Tr \left( (\Svect_v - \Avect_v) (\Svect_w - \Avect_w)^T \right)\): This term captures the pairwise relationships between the consistent graph matrices of different views. The matrices \(\Svect_v\) and \(\Svect_w\) represent the graph matrices for views \(v\) and \(w\), and the term \((\Svect_v - \Avect_v)\) represents the inconsistent part of the graph for view \(v\). Minimizing this term encourages the consistent parts of the graph from different views to be similar to each other. The parameter \(b_{vw}\) controls the weight of the relationship between views \(v\) and \(w\). This term is critical for handling the inconsistencies in the multi-view data and aligning the parts of the data that are inconsistent across views.    
\end{itemize}
This approach includes several constraints to ensure the validity of the graph-based multi-view clustering process. Specifically, the consensus graph matrix \( \Uvect \) must be non-negative, with each row summing to 1. The common spectral representation matrix \( \Fvect \) is required to be orthonormal, meaning \( \Fvect^T \Fvect = \Ivect \), where \( \Ivect \) is the identity matrix. Additionally, the consistent graph matrix \( \Avect_v \) for each view must be non-negative and less than or equal to the graph matrix \( \Svect_v \) of the respective view.}

\added{Together, these terms and constraints aim to learn a consensus graph that captures the most consistent structure across the views, while also considering the individual inconsistencies in each view. The model balances between capturing the global structure (via the consensus graph) and maintaining the consistency within each view (via the consistent graphs). By jointly learning the consensus graph, spectral representation, and the consistent parts of the graphs, this approach enables more accurate and robust multi-view clustering.}

\subsection{Constrained Multi-view Spectral Clustering Via Integrating Nonnegative Embedding and Spectral Embedding (CNESE) }
\label{7-3}

\added{In \cite{Hajjar2021}, the authors address the limitations of spectral clustering-based methods, which usually require an additional clustering step after performing a non-linear projection of the data, which can lead to poorer clustering results due to factors such as initialization or outliers. To overcome these challenges, they propose a constrained version of a method called "Constrained Multi-view spectral clustering via integrating Nonnegative Embedding and Spectral Embedding" (CNESE) by integrating nonnegative embedding and spectral embedding. The model retains the advantages of the original method while incorporating two important constraints: \begin{itemize}
    \item Ensuring consistent smoothing of the nonnegative embedding across all views and,
    \item Imposing an orthogonality constraint on the columns of the nonnegative embedding.
\end{itemize}
This approach provides the clustering result directly and avoids the need to post-process the clustering and use additional parameters by finding the non-negative embedding and the spectral embedding matrix simultaneously.
The proposed objective function aims to jointly estimate the soft cluster assignment matrix $\Hvect$ and the individual spectral projection matrices $\Pvect_v$. This function is given by:}

\begin{equation}
\label{problem}
    \begin{split}
        &\min_{\Pvect_v, \; \Hvect}  \sum_{v=1}^V \, ||\Svect_v \,-\Hvect \, \Pvect_v^T ||_2 \,+ \,\lambda \,\sum_{v=1}^V \,\sqrt{Tr \,\left( \Hvect^T \, \Lvect_v\,\Hvect \right)}
        + \,\alpha \,Tr \,\left(( \Hvect^T \, \Hvect\, - \,\Ivect)^T\,( \Hvect^T \, \Hvect\, - \,\Ivect)   \right) \nonumber\\ 
        &s.t.  \; \;    \Hvect \ge 0,    \; \; \Pvect_v^T \,  \Pvect_v  = \Ivect.
    \end{split}
\end{equation}
where $\lambda$ is a regularization parameter, and $\alpha$ is a large positive value  ensuring the orthogonality of the matrix $\Hvect$ (last term).

\added{The objective function consists of three main terms:
\begin{itemize}
    \item \textbf{Factorization term},
    \(
    \sum_{v=1}^V ||\Svect_v - \Hvect \, \Pvect_v^T ||_2
    \), aims to factorize the input graphs \( \Svect_v \) of each view into the product of the cluster indicator matrix \( \Hvect \) and the view-specific spectral projection matrix \( \Pvect_v \). This term ensures that the clustering information contained in each view is well represented by the cluster assignment and projection matrices.    
    \item \textbf{Smoothing term}, \(
    \sum_{v=1}^V \sqrt{Tr(\Hvect^T \, \Lvect_v \, \Hvect)},
    \) introduces a smoothing effect on the cluster indicator matrix \( \Hvect \). This term promotes consistency of cluster assignments across views by promoting similarity within clusters across views, where \( \Lvect_v \) is a graph Laplacian matrix representing the local structure of the data.
    \item \textbf{Orthogonality constraint}, \(
    \alpha Tr(( \Hvect^T \, \Hvect - \Ivect)^T(\Hvect^T \, \Hvect - \Ivect)),
    \) ensures that the matrix \( \Hvect \) is orthogonal. The orthogonality constraint is controlled by the parameter \( \alpha \), which is a large positive value. This term helps maintain the distinctness of the clusters by enforcing that the cluster indicator matrix \( \Hvect \) forms an orthonormal matrix, meaning that each cluster is represented by a unique vector in the space.
\end{itemize}
Finally, the optimization problem is subject to the following constraints:
\noindent 1. \( \Hvect \ge 0 \), ensuring that the cluster assignment matrix has non-negative entries.
\noindent 2. \( \Pvect_v^T \, \Pvect_v = \Ivect \), ensuring that the projection matrices for each view are orthonormal. 
This objective function jointly optimizes the cluster indicator matrix \( \Hvect \) and the projection matrices \( \Pvect_v \) across all views, taking into account both the consistency within the individual views and the orthogonality of the cluster representations.}

\subsection{ High-Order Multi-View Clustering (HMvC)}
\label{7-4}

\added{In \cite{Pan2023}, the authors propose a novel approach to multi-view clustering, called High-Order Multi-view Clustering (HMvC), which addresses several challenges in graph-based clustering methods. The key points of this approach can be summarized as follows:
\begin{itemize}
 \item \textbf{graph-based clustering with high-order information}: HMvC incorporates higher-order neighborhood information to capture complex interactions within the data that are often overlooked by conventional lower-order methods.
 \item \textbf{Graph filtering for structure encoding}: The approach uses graph filtering to encode structural information and enables unified processing of attributed graph data and non-graph data in a single framework.
 \item \textbf{Exploring long-distance intrinsic relationships}: By utilizing intrinsic relationships up to infinite order, HMvC enriches the learned graph, captures distant connections between data points, and improves the representation of underlying structures.
 \item \textbf{Adaptive graph fusion mechanism}: To integrate the consistent and complementary information from different views, the authors propose an adaptive graph fusion mechanism. This mechanism generates a consensus graph that effectively combines the relevant information from all views.
 \item \textbf{Superior performance}: Experimental results show that HMvC outperforms several state-of-the-art multi-view clustering methods, including some deep learning-based techniques, on both non-graph and attributed graph datasets.
\end{itemize}
This approach highlights the importance of high-order relationships and adaptive fusion mechanisms in achieving robust and effective multi-view clustering. The objective function is given by:}
 
\label{p}
\begin{equation}
\begin{split}
        &\min_{\Svect_v, \; \Svect, \gamma}  \sum_{v=1}^V  \gamma_v\, \left(   ||\Xvect_v^T - \Xvect_v^T \Svect_v ||^2_2 + \alpha\,  || \Svect_v - f_v (\Wvect_v)  ||_2^2 \right) + \beta \,  ||\Svect -   \sum_{v=1}^V \gamma_v\,  \Svect_v||^2_2 + \mu \, ||\Svect ||^2 \\
        &\text{s.t.} \; \; \sum_{v=1}^V   \gamma_v = 1, \gamma_v \ge 0
\end{split}   
\end{equation}
Here the data matrix $\Xvect_v$ corresponds to a filtered version  with  $k$-order filtering based on the adjacency matrix of view $v$, $\Wvect_v$ is a normalized similarity  matrix based on the cosine similarity, and $ f (\Wvect_v) = \Wvect_v + \Wvect^2_v+...+\Wvect^n_v$.

\added{The objective function consists of four main terms:
\begin{itemize}
 \item \textbf{Self-representation term}, \( \sum_{v=1}^V \gamma_v ||\Xvect_v^T - \Xvect_v^T \Svect_v||_2^2,
 \) ensures that the graph of each view \(\Svect_v\) can accurately reconstruct the filtered data matrix \(\Xvect_v^T\). The data matrix \(\Xvect_v\) is derived by filtering the order \(k\) based on the adjacency matrix of the view \(v\), which captures the structural information within the view.
 \item \textbf{Higher- order consistency term}, \(
 \sum_{v=1}^V \gamma_v \alpha ||\Svect_v - f_v(\Wvect_v)||_2^2,
 \) includes the higher order neighborhood information in the graph construction. Here, \(f_v(\Wvect_v) = \Wvect_v + \Wvect_v^2 + \cdots + \Wvect_v^n\) aggregates the information from multi-order relationships based on the normalized similarity matrix \(\Wvect_v\). This term ensures that the learned graph \(\Svect_v\) matches the higher order structural information within each view.
 \item \textbf{Consensus graph alignment term}, \(
 \beta ||\Svect - \sum_{v=1}^V \gamma_v \Svect_v||_2^2,
 \) encourages the consensus graph \(\Svect\) to be a weighted combination of the view-specific graphs \(\Svect_v\). The weights \(\gamma_v\) are adaptively learned to reflect the importance of each view and to ensure consistency between views.
 \item \textbf{Regularization term}, \(
 \mu ||\Svect||^2,
 \) serves as a regularization of the consensus graph \(\Svect\) to promote smoothness and avoid overfitting.
\end{itemize}
The constraints \(\sum_{v=1}^V \gamma_v = 1\) and \(\gamma_v \geq 0\) ensure that the weights \(\gamma_v\) form a valid convex combination that maintains the balance between the contributions of the different views}

\added{This formulation effectively integrates self-representation, higher-order relations, and adaptive graph fusion, making it robust for multiview clustering tasks across different datasets.}

\subsection{Multi-view Structured Graph Learning  (MSGL) }
\label{7-5}

\added{In \cite{Kang2022}, the authors propose a scalable graph learning framework tailored for subspace clustering that addresses three key challenges of existing graph-based methods: high computational cost, the inability to explicitly detect clusters, and the lack of generalizability to unseen data. The main contributions of their work are summarized as follows:
\begin{itemize}
 \item \textbf{Bipartite graph construction}: Instead of constructing a full graph for \(n\) samples, the proposed framework builds a bipartite graph to model the relationships between data samples and anchor points. This approach significantly reduces computational complexity and improves scalability, making the method suitable for large datasets.
 \item \textbf{cluster interpretability}: The method contains a connectivity constraint that ensures that the connected components in the bipartite graph correspond directly to the clusters. This eliminates the need for an additional clustering step and the cluster memberships are explicitly displayed.
 \item \textbf{link to K-means clustering}: The authors provide a theoretical link between their learning approach for bipartite graphs and the K-means clustering algorithm, which provides insights into its clustering mechanism and further justifies its effectiveness.
 \item \textbf{extension to multi-view data}: The framework is extended to handle multi-view datasets. This multi-view model achieves linear scalability with respect to \(n\), maintaining computational efficiency while effectively integrating information from multiple views.
\end{itemize}
The work in \cite{Kang2022} represents a significant advancement in graph-based clustering, particularly for large-scale and multi-view datasets, by combining efficiency, explicit cluster discovery, and the ability to generalize to new data points.
The objective function in \cite{Kang2022} is designed to tackle the challenges of scalability, explicit cluster identification, and multi-view data integration in subspace clustering. The function is expressed as:}

\begin{align}
&\min_{\alpha,   \Zvect, \Pvect} \sum_{v=1}^V  \alpha_v \, ||\Xvect_v - \Avect_v \,  \Zvect^T||^2_2 + \lambda_1 ||  \Zvect ||^2_2  + \lambda_2  \, Tr \, (\Pvect^{T} \, \Lvect \, \Pvect) + \sum_{v=1}^V \alpha_v ^{\gamma} \nonumber\\ 
&s.t.  \; \;   \Pvect^T \,  \Pvect = \Ivect,  \; \; \Zvect \ge 0,    \; \; \Zvect^T \,  \unvect = \unvect, \gamma < 0 
\end{align}
where $\Avect_v$ is the matrix of anchors in view $v$, $\Zvect$ is the consensus graph matrix between data and their anchors. 

{\added{The terms and constraints in this objective function are described below:
\begin{itemize}
    \item \textbf{Reconstruction term},
    \(
    \sum_{v=1}^V \alpha_v \, ||\Xvect_v - \Avect_v \,  \Zvect^T||^2_2,
    \)
    models the reconstruction error between the data matrix \(\Xvect_v\) of view \(v\) and the product of the anchor matrix \(\Avect_v\) and the consensus graph matrix \(\Zvect^T\). This term ensures that the learned bipartite graph accurately represents the relationship between the data points and their anchors, weighted by the view-specific coefficient \(\alpha_v\).
    \item \textbf{Consensus graph regularization},
    \(
    \lambda_1 ||  \Zvect ||^2_2,
    \)
    imposes an \(\ell_2\)-norm penalty on the consensus graph matrix \(\Zvect\), encouraging sparsity and regularizing the learned graph.
    \item \textbf{Smoothness term},
    \(
    \lambda_2  \, Tr \, (\Pvect^{T} \, \Lvect \, \Pvect),
    \)
    promotes the smoothness of the clustering solution by incorporating the Laplacian matrix \(\Lvect\). Here, \(\Pvect\) is the projection matrix encoding the cluster memberships, and this term ensures that points connected in the graph are assigned similar cluster memberships.
    \item \textbf{Weight regularization term},
    \(
    \sum_{v=1}^V \alpha_v ^{\gamma},
    \)
    regularizes the view-specific weights \(\alpha_v\), controlling their influence on the clustering process. The hyperparameter \(\gamma < 0\) promotes a balanced contribution from different views.
\end{itemize}
The constraints in the objective function serve to ensure the validity and interpretability of the learned matrices. The orthogonality constraint \(\Pvect^T \, \Pvect = \Ivect\) guarantees that the projection matrix \(\Pvect\) encodes unique cluster memberships, where the clusters are represented as distinct, non-overlapping components. }

\added{The non-negativity condition \(\Zvect \ge 0\) enforces that the consensus graph matrix \(\Zvect\) contains only non-negative entries, which is crucial for interpretability as it represents the relationships between data points and anchor points in a meaningful way. }

\added{Finally, the normalization constraint \(\Zvect^T \, \unvect = \unvect\) ensures that each row of \(\Zvect\) sums to one, effectively treating it as a valid probability distribution. This combination of constraints ensures a robust and meaningful clustering solution.
This objective function integrates graph learning and clustering in a unified framework, and ensures scalability, interpretability and robustness in both single-view and multi-view subspace clustering scenarios.
}

\subsection{Fast Parameter-Free Multi-view Subspace Clustering with Consensus Anchor Guidance (FPMVS-CAG)}
\label{7-6}

\added{In \cite{Wang-Siwei2022}, the authors propose a new method called Fast Parameter-free Multi-view Subspace Clustering with Consensus Anchor Guidance (FPMVS-CAG). This approach addresses the main limitations of existing multi-view subspace clustering techniques, especially their cubic time complexity and dependence on heuristic anchor sampling strategies. The main contributions of this method can be summarized as follows:
\begin{itemize}
 \item \textbf{Unified Optimization Framework}: FPMVS-CAG integrates anchor selection and subspace graph construction into a single optimization problem. This joint formulation allows the two processes to interact and reinforce each other, improving the quality of clustering.
 \item \textbf{parameter-free learning}: Unlike conventional methods that require manual tuning of hyperparameters, FPMVS-CAG automatically learns an optimal anchor subspace graph without introducing additional parameters. This eliminates the time-consuming selection of parameters, which improves the applicability of the method.
 \item \textbf{scalability}: The method achieves linear time complexity with respect to the number of samples. This makes it particularly suitable for large-scale applications and eliminates the inefficiency of previous methods with cubic complexity.
\end{itemize}
The combination of these properties ensures that FPMVS-CAG is well suited for large-scale multi-view clustering tasks while maintaining high accuracy and robustness}

To achieve this, the authors proposes to estimate a graph matrix that relate the data to their anchors, optimizing the following objective function:
\begin{align}
&\min_{\alpha, \Wvect_v=1,...,V,  \Avect, \Zvect} \sum_{v=1}^V  \alpha_v^2 \, ||\Xvect_v - \Wvect_v \, \Avect \,  \Zvect||^2_2  \nonumber\\ 
&s.t.  \; \;  \Wvect_v^T \,  \Wvect_v  = \Ivect, \Avect^T \,  \Avect = \Ivect,  \; \; \Zvect \ge 0,    \; \; \Zvect^T \,  \unvect = \unvect, \; \; \alpha^T \, \unvect =1 
\end{align}
where ${\Xvect_v} \in \mathbb{R}^{d_{v}\times n}$  is the data matrix in view $v$,  the respective projection matrices $\Wvect_v, v=1,...,V$
target the consensus anchor guidance, $\Avect \in \mathbb{R}^{d \times l}$ is the latent consensus anchor matrix (a learnable dictionary matrix). The number and size of the anchors, $d$ and $l$ are specified beforehand).

\added{The objective function aims to minimize the discrepancy between the data matrices and their reconstruction across all views. This term enforces that the data from each view is well represented by the anchor and projection matrices, ensuring consistency between multiple views.}

\added{The constraints ensure meaningful representations: The projection matrices \( \Wvect_v \) and the anchor matrix \( \Avect \) are orthonormal. The consensus graph matrix \( \Zvect \) must be non-negative (\( \Zvect \geq 0 \)) and its rows are normalized such that \( \Zvect^T \unvect = \unvect \), where \( \unvect \) is a vector of ones. In addition, the weights \( \alpha_v \) are normalized, where \( \alpha^T \unvect = 1 \).}

\added{Once the consensus graph matrix \( \Zvect \) is obtained, its singular vectors are used as the consensus spectral embedding. The final clustering is achieved by applying the \( k \)-mean algorithm to these embeddings. This framework integrates the reconstruction of data from multiple views, anchor learning and spectral clustering into a unified model.}

\subsection{
Efficient Orthogonal Multi-view Subspace Clustering
(OMSC)}
\label{7-7}

\added{In \cite{Chen-Man2022}, the authors propose a method titled Efficient Orthogonal Multi-view Subspace Clustering (OMSC), which addresses the challenges of clustering multi-view data in large-scale scenarios. The main contributions can be outlined as follows:
\begin{itemize}
 \item \textbf{Integrated approach}: OMSC combines anchor selection, graph construction and clustering in a cohesive framework. This integration ensures that these components reinforce each other, resulting in more robust and adaptive anchor representations and cluster assignments.
 \item \textbf{Learning Orthogonal Bases}: In contrast to traditional approaches that predefine anchors using \( k \)-means or uniform sampling, OMSC introduces a mechanism to learn high-quality orthogonal bases within the unified model. This approach increases the algebraic structure and improves the accuracy of clustering.
 \item \textbf{High scalability}: The model achieves near linear complexity with respect to the size of the dataset, making it suitable for clustering tasks with extremely large datasets. This efficiency results from the joint modeling process and a carefully developed alternative optimization strategy.
\end{itemize}
By addressing these critical aspects, OMSC provides a scalable and effective solution for clustering multi-view data in real-world, large-scale applications \cite{Chen-Man2022}}. 
The objective function is given by:
\begin{align}
\label{eq-7-7}
&\min_{\alpha, \Wvect_v,  \Avect, \Zvect, \Gvect, \Fvect} \sum_{v=1}^V  \alpha_v^2 \, ||\Xvect_v - \Wvect_v \, \Avect \,  \Zvect||^2_2 + \lambda \, ||\Zvect - \Gvect \, \Fvect||^2_2  \nonumber\\ 
 &s.t.  \; \;  \Wvect_v^T \,  \Wvect_v  = \Ivect, \Avect^T \,  \Avect = \Ivect,  \; \; \Zvect \ge 0,    \; \; \Zvect^T \,  \unvect = \unvect, \; \; \alpha^T \, \unvect = 1,  \nonumber\\
 &\Gvect^T \,  \Gvect  = \Ivect,  F_{ij}  \in \{0, 1\}, \sum_{i=1}^{k} F_{ij} =1
\end{align}
where $\Avect \in \mathbb{R}^{d \times l}$ is the latent consensus anchor matrix, $\Zvect$ is the consensus graph matrix between data and their anchors,  $\Gvect \in \mathbb{R}^{l \times k} $ is the centroid matrix ($k$ being the number of clusters) and $\Fvect \in \mathbb{R}^{k \times n}$ is an indicator matrix providing the data partition, $F_{ij} = 1$ if the i-th instance is
assigned to the $k$-th cluster and 0 otherwise. 
\added{
\begin{itemize}
 \item The first term, \(\sum_{v=1}^V \alpha_v^2 \, ||\Xvect_v - \Wvect_v \, \Avect \, \Zvect||^2_2\), minimizes the reconstruction error for each view \(v\). Where \(\Xvect_v\) is the data matrix for the view \(v\), \(\Wvect_v\) is the projection matrix, \(\Avect\) is the latent consensus anchor matrix and \(\Zvect\) is the consensus graph matrix. The weights \(\alpha_v^2\) balance the contribution of the individual views.
 \item The second term, \(\lambda \, ||\Zvect - \Gvect \, \Fvect||^2_2\), ensures that the consensus graph matrix \(\Zvect\) matches the cluster representation formed by the centroid matrix \(\Gvect\) and the cluster indicator matrix \(\Fvect\). The parameter \(\lambda\) controls the strength of the regularization.
\end{itemize}
}

\subsection{Anchor-Based Multi-View Subspace
Clustering with Graph Learning (AMVSCGL)}
\label{7-8}

\added{The article \cite{Su2023} introduces a novel method to address the challenges in Multi-view Subspace Clustering (MVSC), a key problem in pattern recognition and data mining. This method generates a joint coefficient matrix rather than constructing large, view-specific graphs, which enhances scalability and reduces memory consumption. Additionally, it incorporates a graph learning term that combines both global and local information from multiple views, making it more efficient and adaptable for large datasets. In contrast, \cite{Chen-Man2022} relies on predefined orthogonal anchors and constructs affinity graphs for each view, which can be computationally expensive and less scalable for large datasets. Furthermore, \cite{Su2023} is parameter-free, making it easier to implement, while \cite{Chen-Man2022} requires the manual tuning of hyperparameters, such as the number of anchors.}

The objective function is as follows.

\begin{align}
\min_{\alpha, \Avect_v,  \Zvect} \sum_{v=1}^V  \left ( \alpha_v\, ||\Xvect_v - \Avect_v \,  \Zvect^T||^2_2 + \lambda  \sum_{i=1}^n \sum_{j=1}^k  ||  \Xvect_v (i, :) - \Avect_v (:, j)  ||^2_2  + \alpha_v ^{\gamma}  \right ) \nonumber\\ 
 s.t.  \; \;   \Avect_v^T \,  \Avect_v = \Ivect_k,  \; \; \Zvect \ge 0,    \; \; \Zvect^T \,  \unvect = \unvect, \gamma < 0 &
\end{align}
where $k$ is the number of clusters, $\Avect_v$ is the matrix of anchors in view $v$ (a learnable matrix), $\Zvect$ is the consensus graph matrix between data and their anchors. 
\added{ 
\begin{itemize}
    \item \textbf{First term:} \( \alpha_v\, ||\Xvect_v - \Avect_v \,  \Zvect^T||^2_2 \). This term measures the reconstruction error between the data in view \( v \) (\( \Xvect_v \)) and the approximation of the data using the anchor matrix \( \Avect_v \) and the consensus graph matrix \( \Zvect \). The weight \( \alpha_v \) controls the contribution of each view to the overall objective.
    \item \textbf{Second term:} \( \lambda \sum_{i=1}^n \sum_{j=1}^k || \Xvect_v (i, :) - \Avect_v (:, j) ||^2_2 \). This term ensures that the data points \( \Xvect_v(i, :) \) are well approximated by the anchors \( \Avect_v(:, j) \) from the corresponding view. The regularization parameter \( \lambda \) controls the strength of this term and helps to maintain the consistency of the anchors across the different views.
    \item \textbf{Third term:} \( \alpha_v^{\gamma} \). This term introduces a penalty based on the value of the weight \( \alpha_v \) for each view. The exponent \( \gamma \) is a negative constant, which usually leads to a regularization effect that favors smaller values of \( \alpha_v \). This helps to control the influence of the individual views on the overall model so that the method can handle different contributions of the views more flexibly.
\end{itemize}
}

\subsection{Self-Taught Multi-View Spectral Clustering (SMSC)}
\label{7-9}

\added{The work \cite{Zhong-Guo2023} focuses on the relax-and-discretize strategy, which uses predefined similarity graphs and learns a consensual Laplacian embedding for clustering. The main contributions of this work are as follows:
\begin{itemize}
 \item \textbf{Problem addressed:} The work addresses the problem of information loss in MVC methods that arises from the independent processing of similarity graphs and the subsequent graph partitioning steps. This approach is often inefficient and reduces the quality of clustering.
 \item \textbf{Proposed framework:} The authors present the \textit{Self-taught Multi-view Spectral Clustering (SMSC)} framework. This method considers both the manifold structure induced by Laplacian embeddings and the cluster information embedded in the discrete indicator matrix, which enables learning an optimal consensus similarity graph.
 \item \textbf{Graph Fusion Schemes:} Two graph fusion schemes are presented:
 \begin{itemize}
 \item \textbf{Convex combination scheme:} This approach combines the similarity graphs from different views by a convex combination.
 \item \textbf{Centroid Graph Fusion Scheme:} In this method, the similarity graphs are merged taking into account the centroid of each view's data representation of the individual views.
 \end{itemize}
 \item \textbf{Self-taught Mechanism:} The self-taught mechanism integrates manifold structure and clustering information to learn an optimal consensus similarity graph for graph partitioning and thus improve clustering results.
\end{itemize}
These contributions make SMSC a powerful and efficient approach by overcoming the challenges of information loss and optimizing the learning process for clustering from multiple views.} The objective function of this approach is given by:
\begin{align}
\label{eq-7-9}
\min_{\alpha,  \Svect,  \Pvect, \Yvect}   || \sum_{v=1}^V  \alpha_v \,  \Svect_v -  \Svect||^2_2 + \lambda_1  \, Tr \, (\Pvect^{T} \, \Lvect \, \Pvect) + \lambda_2  \, ||\Pvect - \Yvect \, (\Yvect^T \Yvect)^{-1}||^2_2  \nonumber\\ 
 s.t.  \; \;  \Pvect^T \,  \Pvect  = \Ivect, \; \; \Svect \ge 0,  \Svect^T \,  \unvect = \unvect,  \; \; \alpha \,  \unvect = 1, \;\;
  Y_{ij}  \in \{0, 1\}, \sum_{i=1}^{k} Y_{ij} =1&
\end{align}
where $\Svect$ denotes the consensus graph matrix, 
 $\Pvect$ denotes the consensus spectral representation matrix, 
where $\Yvect$ denotes the indicator matrix. The terms that constitute the objective function, as given by Eq. \ref{eq-7-9}, are described as follows:

\added{\begin{itemize}
 \item \textbf{The first term},
 \(\, || \sum_{v=1}^V \alpha_v \, \Svect_v - \Svect||^2_2\), penalizes the difference between a weighted sum of view-specific matrices \( \Svect_v \) and the consensus matrix \( \Svect \). The weights for each view are labeled \( \alpha_v \), and the target encourages the consensus matrix \( \Svect \) to be close to the weighted sum of view-specific matrices.
 \item \textbf{The second Term}, 
 \(\, \lambda_1 \, Tr \, (\Pvect^{T} \, \Lvect \, \Pvect)\), 
 regularizes the projection matrix \( \Pvect \) by encouraging it to align with the Laplacian of the graph \( \Lvect \), where \( \lambda_1 \) is a regularization parameter. This ensures that the data points are mapped in such a way that the structure defined by the Laplacian matrix is preserved.
 \item \textbf{The third term}, 
 \(\, \lambda_2 \, ||\Pvect - \Yvect \, (\Yvect^T \Yvect)^{-1}||^2_2\), 
 enforces consistency between the projection matrix \( \Pvect \) and a matrix derived from \( \Yvect \), where \( \Yvect \) is an indicator matrix that provides the cluster assignments. The regularization parameter \( \lambda_2 \) controls the weighting of this term and promotes the alignment between \( \Pvect \) and the cluster structure represented by \( \Yvect \).
\end{itemize}}

The authors also propose another learning model that releases the estimation of the blending weights $\alpha_v$. They propose to minimize the following objective function with the concept of automatic view weighting:
\begin{align}
\min_{\alpha,  \Svect,  \Pvect, \Yvect}    \sum_{v=1}^V    || \Svect_v -  \Svect||^2_2 + \lambda_1  \, Tr \, (\Pvect^{T} \, \Lvect \, \Pvect) + \lambda_2  \, ||\Pvect - \Yvect \, (\Yvect^T \Yvect)^{-1}||^2_2  \nonumber\\ 
 s.t.  \; \;  \Pvect^T \,  \Pvect  = \Ivect, \; \; \Svect \ge 0,    \; \; \Svect^T \,  \unvect =  \unvect, \;\;
  Y_{ij}  \in \{0, 1\}, \sum_{i=1}^{k} Y_{ij} =1&
\end{align}

\subsection{
Multi-view Clustering Using Unified Graph Learning and Spectral Representation (MCGLSR)}
\label{7-10}
\modified{Most existing multi-view clustering methods follow a sequential process in three steps: estimating individual or consistent similarity matrices, performing spectral embedding, and then partitioning into clusters. These methods often reach their limits when integrating all the necessary components. To address this issue, the authors in \cite {Dornaika2023} present a novel approach to multi-view clustering which integrates multiple components into a single framework, overcoming the weaknesses of previous methods. It jointly solves the consistent similarity matrix for all views, the spectral representation, the soft cluster assignments and the weights of the views within a single criterion
The key innovations are as follows:
 \begin{itemize}
 \item The method eliminates the need for an additional clustering step as it resolves the clustering assignments directly.
 \item The soft cluster assignments are directly linked to the representation of the views, which improves consistency across all views.
 \end{itemize}
This approach is characterized by its ability to jointly optimize multiple components of the clustering process without the need for an additional partitioning step. This provides a more efficient and integrated solution compared to conventional methods.}

The proposed objective function is given by:
 \begin{align}
 \label{eq-7-10}
 \min_{\Svect^v,  \Svect^*,  \Pvect^*, \Hvect} \sum_{v=1}^V \left \{  Tr \, (\Kvect^v\,-2 \, \Kvect^v \, \Svect^v +\Svect^{vT} \, \Kvect^v \, \Svect^v )  +  \,||\Svect^v||^{2}_2\,+  \, \lambda_1 \, ||\Svect^* -\, \Svect^v ||^{2}_2 \right \} \,  \nonumber\\  
 +\, \lambda_2 \, ||\Svect^* -\Hvect \,  \Pvect^{*T} ||^{2}_2 \, 
 + \, 
 \lambda_3 \, Tr \, (\Hvect^T \, \Lvect^* \, \Hvect) \, 
 + \, \lambda_4 \, Tr \, (\Pvect^{*T} \, \Lvect^* \, \Pvect^*), &
  \end{align}
where  $\lambda_1$, $\lambda_2$ and $\lambda_3$ and $\lambda_4$ are regularization parameters.  ${\Kvect_v} \in \mathbb{R}^{n\times n}$ is the kernel data matrix in view $v$,   ${\Svect_v} \in \mathbb{R}^{n\times n}$  is the graph matrix in view $v$,  ${\Svect^*} \in \mathbb{R}^{n\times n}$ the consensus graph matrix,   $\Hvect \in \mathbb{R}^{n \times k}$ is the matrix of soft cluster assignments, and  ${\Pvect^*} \in \mathbb{R}^{n\times n}$ is the consensus spectral representation. The terms that constitute the objective function, as given by Eq. \ref{eq-7-10}, are described as follows:

\modified{\begin{itemize}
 \item \textbf{First term (for each view \( v \)):}
 \(\text{Tr} \, (\Kvect^v - 2 \, \Kvect^v \, \Svect^v + \Svect^{vT} \, \Kvect^v \, \Svect^v)\)
 This term calculates the trace of the difference between the kernel matrix \(\Kvect^v\) and the graph matrix \(\Svect^v\), which measures the alignment between the data and the graph for each view.
 \item \textbf{Second term (graph matrix regularization):}
 \(||\Svect^v||_2^2\)
 This regularization term applies the \(\ell_2\) norm to the graph matrix \(\Svect^v\), which promotes sparsity and reduces overfitting.
 \item \textbf{Third term (consistency between graphs):}
 \(\lambda_1 ||\Svect^* - \Svect^v||_2^2\)
 This term ensures that the consensus graph matrix \(\Svect^*\) is similar to each individual graph matrix \(\Svect^v\), enforcing consistency between the different views.
 \item \textbf{Fourth term (regularization of the consensus representation):}
 \(\lambda_2 ||\Svect^* - \Hvect \, \Pvect^{*T}||_2^2\)
 This term forces the consensus graph matrix \(\Svect^*\) to be close to the product of the soft cluster assignments \(\Hvect\) and the consensus spectral representation \(\Pvect^{*T}\), which strengthens the relationship between the graph and the clustering.
 \item \textbf{Fifth term (smoothness in the consensus representation):}
 \(\lambda_3 \, \text{Tr} \, (\Hvect^T \, \Lvect^* \, \Hvect)\)
 This term regulates the cluster assignment matrix \(\Hvect\) by ensuring a smoothing of the cluster labels via the Laplacian matrix \(\Lvect^*\).
 \item \textbf{Sixth term (regularization of the spectral representation):}
 \(\lambda_4 \, \text{Tr} \, (\Pvect^{*T} \, \Lvect^* \, \Pvect^*)\)
 This term regulates the consensus spectral representation \(\Pvect^*\) by ensuring smoothing over the Laplacian matrix \(\Lvect^*\) and preserving the structure of the data across the views.
\end{itemize}}

\section{Dataset Exploration}
\label{datasets}

In this section, we provide a comprehensive overview of the key datasets that are important for the evaluation of multi-view clustering methods. We provide relevant information such as the number of samples, the number of views and the number of clusters. Table \ref{tab:dataset-info} gives an overview of the common datasets including image-based and feature-based datasets.

Indeed, features are usually extracted either by deep neural networks or manually created features. The choice of feature extraction method plays a crucial role in the effectiveness of the clustering process. 

Deep neural networks have proven to be powerful tools for feature extraction in various domains due to their ability to automatically learn hierarchical representations from raw data. Notable architectures such as convolutional neural networks (CNNs) and recurrent neural networks (RNNs) have been extensively used for extracting discriminative features from images in multi-view clustering scenarios (\cite{lecun2015deep, goodfellow2016deep}). These networks show that they are able to capture complicated patterns and relationships within the data and thereby improve the quality of the extracted features for subsequent clustering tasks.

In contrast, handcrafted features \cite{moujahid2020multi,moujahid2019pyramid,lan2015quaternionic,tan2010enhanced} involve a careful design process that utilizes expertise to develop features that contain relevant information for clustering purposes. The handcrafted feature selection process requires a sophisticated understanding of the characteristics of the dataset, and the features chosen should encapsulate the inherent structures of the data to enable effective clustering.

It is important to note that the number of views within a given dataset is a variable parameter that varies in different studies and applications. Determining the optimal number of views is a crucial consideration that affects the overall performance of the multi-view clustering algorithm. Researchers have applied various strategies to solve this problem. They range from empirical selection based on expert knowledge to automated methods guided by statistical criteria \cite{LIN2023110954,Chao2021}.

  \begin{longtable}{|p{2.5cm}|p{8cm}|p{1.2cm}|p{1cm}|p{1.2cm}|}
\caption{Exploring Variability in Multi-View Clustering Datasets.} 
\label{tab:dataset-info}\\
\hline
\textbf{Dataset} & \textbf{Description} & \textbf{Samples} & \textbf{Views} & \textbf{Clusters} \\
\hline
\endfirsthead

\multicolumn{4}{c}%
{{\tablename\ \thetable{} -- Continued from previous page}} \\
\hline
\textbf{Dataset} & \textbf{Description} & \textbf{Samples} & \textbf{Views} & \textbf{Clusters} \\
\hline
\endhead

\hline \multicolumn{4}{|r|}{{Continued on next page}} \\ \hline
\endfoot

\hline
\endlastfoot

    \hline
    Caltech101 \cite{Li2004}\footnote{https://data.caltech.edu/records/mzrjq-6wc02} & The Caltech101 dataset contains images of objects belonging to 101 categories. Here are two of the most frequently used subcategories:  & 	&  &  \\
     & - Caltech101-7 is a subset with 7 object categories  & 1474	& 6 & 7 \\    
     & - Caltech101-20 is a subset with 20 object categories & 2386 & 6 & 20 \\
    \hline
    SUN-RGBD \cite{song2015sun} & Large-scale indoor scene understanding dataset with RGB-D images & 10335 & 45 & 2 \\
    \hline
    Animal \cite{lampert2013attribute}  & Subset of the Animals with Attributes dataset (AwA) dataset containing images of animals with attribute annotations & 11673	& 20 & 4 \\
    \hline
    AWA \cite{lampert2013attribute}   & The Animals with Attributes dataset (AwA) containing images of animals with attribute annotations & 30475 & 50 & 6 \\
    \hline    
    NUS-WIDE \cite{chua2009nus}  & Dataset with web images and associated object labels & 30000 & 31 & 5 \\
    \hline
    YoutubeFace\footnote{https://www.cs.tau.ac.il/ wolf/ytfaces/}  & Dataset focused on face recognition with video frames and face labels & 101499 & 31 & 5 \\
    \hline
    BBC-Sport\footnote{http://mlg.ucd.ie/datasets/bbc.html}  & Collection of articles from the sports section of the BBC website relating to sports news in five subject areas, including athletics, cricket, football, rugby, and tennis. Each article has two views \cite{Li2024}. The first view consists of 3183 features, and the second view consists of 3203 features. & 544 & 2 & 5 \\
    \hline
    COIL-20\footnote{https://www.cs.columbia.edu/CAVE/software/softlib/coil-20.php}  & Contains 1,440 grayscale images, divided into 20 classes, each class consisting of 72 images. Each image was taken from three different  \cite{Dornaika2023}:  intensity feature (1024), LBP feature (3304) and Gabor feature (6750). & 1440 & 3 & 20 \\
    \hline
    UCI-digits\footnote{http://archive.ics.uci.edu/ml/datasets/Multiple+Features}  & Comprises 2,000 data points divided into 10 digit classes, each class containing 200 handwritten digits. The dataset includes 3 views \cite{LuYL15}: profile correlation (216), Fourier coefficient (76) and Karhunen-Loeve coefficient (64) & 2000 & 6 & 10 \\
    \hline
    ORL\footnote{http://cam-orl.co.uk/facedatabase.html/}  & Consists of 40 individuals, with each individual represented by 10 different images. Each image is characterized by three views \cite{Li2024}:  intensity feature (4096), LBP feature (3304) and Gabor feature (6750). & 400 & 3 & 40 \\
    \hline   
    Outdoor Scene  & Consist of 2688 images. These images are divided into 8 groups. Each image is characterized by 4  \cite{monadjemi2002experiments}: GIST features (512), color moment features (432), HOG features (256), and LBP features (48). & 2688 & 4 & 8 \\
    \hline    
    MSRCv1  &  Consists of 210 instances from Microsoft Research in Cambridge divided into 7 groups. Each image is characterized by 5 views \cite{winn2005locus} :  GIST features (512), color moment features (24), CENTRIST features (254), SIFT features (512), and LBP features (256). & 210 & 5 & 7 \\
    \hline  
    COVIDx\footnote{https://github.com/lindawangg/COVID-Net/blob/master/docs/COVIDx.md}  &  Consists of 13892 samples divided into 5458 instances corresponding to the pneumonia class, 468 instances corresponding to the COVID19 class, and 7966 instances corresponding to the normal class. Each image is characterized by 3 views:  ResNet50 (2048), ResNet101 (2048), and DenseNet169 feature vector (1664). & 13892 & 3 & 2 \\
    \hline    
\end{longtable}

\section{Conclusion}
In this comprehensive survey we systematically explored the landscape of multi-view clustering, providing a structured and insightful journey through various aspects of the field. We started with an introduction where we highlighted the increasing importance of multi-view clustering in modern machine learning and data analytics and explained both the challenges and motivations. The following sections covered the basics of multi-view clustering, classical methods, kernel-based approaches, subspace-based methods and depth-based techniques, providing a differentiated understanding of each method.

We also looked at graph-based multi-view clustering and introduced novel algorithms that can overcome specific challenges of multi-view clustering such as missing data, noise suppression and computational efficiency. The section on multi-view clustering with missing or incomplete data looked at recent developments in data analysis, focusing on the challenges posed by data incompleteness in clustering scenarios.

A careful examination of typical approaches was conducted in a formal mathematical sense, shedding light on the mathematical framework underlying these methods. This rigorous analysis contributes to a deeper understanding of the methods used in multi-view clustering and focuses on the formal aspects of their mathematical foundations.

In addition, a comprehensive overview of the key datasets that are critical to the evaluation of multi-view clustering methods is provided. This includes a concise summary of essential details covering common feature extraction methods and a critical examination of how to determine the optimal number of views.

To conclude, this review not only summarizes existing knowledge but also lays the foundation for future advances in multiview clustering research. By systematically organizing and presenting the various facets of the field, it serves as a valuable resource for researchers seeking to foster a deeper understanding of the complexity and potential of the multiview clustering landscape. By the end of this overview, it is clear that multi-view clustering is a dynamic and evolving field with a variety of perspectives.

\section{Future Directions}
\label{future-directions}

\added{Multi-view clustering has achieved significant progress in overcoming various challenges related to the integration of different data views. These advances have laid the foundation for robust methods and various applications, but much remains to be done to tackle unsolved problems and exploit new opportunities. The following key areas represent both milestones achieved and exciting prospects for future research:
\begin{itemize}
\item \textbf{Integration with deep learning:}
Multi-view clustering has already benefited significantly from the integration of deep learning techniques. Methods such as autoencoders and graph neural networks have enabled more flexible and accurate learning of representations \cite{Zhu2024}. However, challenges such as the need for effective unsupervised architectures and scalable models remain. Future research should focus on developing architectures that exploit correlations between multiple views, taking into account advances in unsupervised deep learning and self-supervised paradigms.
\item \textbf{Adaptive and self-learning mechanisms:}
Recent developments have introduced adaptive models that are able to dynamically weight views based on their contributions, improving robustness to noisy or incomplete views \cite{Wang-Siwei2022}. However, these methods often rely on hand-crafted assumptions. Future work could focus on self-learning frameworks that are able to automatically discover optimal representations and weighting strategies, and leverage advances in meta-learning and reinforcement learning to refine their adaptability.
\item \textbf{Scalability to large data sets:}
Progress has already been made in scaling multi-view clustering algorithms using techniques such as anchor-based methods and distributed computing systems \cite{Wang-Siwei2022,Chen-Man2022}. However, the exponential growth in data volume and complexity requires further innovation. Research needs to focus on lightweight algorithms with linear or sublinear complexity that are capable of processing streaming or real-time multi-view data in distributed environments.
\item \textbf{Heterogeneity of views:}
Existing multi-view clustering methods have addressed heterogeneity by using techniques such as kernel-based approaches and joint latent space learning \cite{Li-Kai2022}. Despite these efforts, dealing with different distributions, dimensions, and noise levels between views remains a challenge. Future research should explore robust frameworks that can dynamically adapt to heterogeneous data features while ensuring interpretability and effectiveness.
\item \textbf{Interpretability and explainability:}
Multi-view clustering models increasingly use latent representation learning to improve performance, but this often reduces transparency. To improve interpretability, frameworks such as Self-taught Multi-view Spectral Clustering (SMSC) \cite{Zhong-Guo2023} integrate manifold structure with clustering information to create a consensus similarity graph that matches the cluster representations with the input graphs. Similarly, Multi-view Clustering using Unified Graph Learning and Spectral Representation (MCGLSR) \cite{Dornaika2023} ensures consistency between views while resolving cluster mappings directly. Despite these advances, it remains a challenge to develop fully interpretable multi-view clustering models.
Future directions should focus on developing models that not only provide accurate clustering results, but also provide actionable insights into the relationships between the views and the clustered data.
\end{itemize}
Significant progress has been made in overcoming the main challenges of multi-view clustering, such as data integration and scalability. However, further progress in the areas of interpretability, heterogeneity and adaptive learning are crucial for further development. The integration of emerging trends such as deep learning, explainable AI and domain-specific adaptation will play a crucial role in improving the relevance and robustness of multi-view clustering methods for tackling complex real-world problems.}

\added{In recent years, domain-specific knowledge, especially in areas such as bioinformatics and medical diagnostics, has been successfully applied to improve clustering performance \cite{benton-etal-2019-deep, Feng2022}. Nevertheless, frameworks for the seamless integration of domain-specific priorities into clustering algorithms have not yet been sufficiently explored. Future research could focus on developing automated approaches to encode domain knowledge using tools such as knowledge graphs and ontology-based constraints to improve the accuracy and applicability of clustering.}

\end{document}